\documentclass[10pt,twocolumn,letterpaper]{article}

\usepackage{iccv}              

\usepackage{graphicx}
\graphicspath{{/images/}}
\usepackage{amsmath}
\usepackage{amssymb}
\usepackage{booktabs}
\usepackage{multicol}
\usepackage{multirow}
\usepackage{algorithm}
\usepackage{algpseudocode}
\usepackage{comment}
\usepackage{colortbl}
\usepackage{array}
\usepackage[inkscape]{svg}
\usepackage{placeins}
\definecolor{LightYellow}{RGB}{255, 249, 196}
\definecolor{LightGray}{gray}{0.92}
\definecolor{RoyalBlue}{RGB}{65,105,225}
\definecolor{Emerald}{RGB}{80,200,120}
\definecolor{DeepPink}{RGB}{255,20,147}
\definecolor{Yellow}{RGB}{255,255,0}
\definecolor{customYellow}{HTML}{FFD966}
\definecolor{customGreen}{rgb}{0.88, 1, 0.88} 

\usepackage{subcaption}  
\usepackage[accsupp]{axessibility}

\definecolor{iccvblue}{rgb}{0.21,0.49,0.74}
\usepackage[pagebackref,breaklinks,colorlinks,allcolors=iccvblue]{hyperref}
\usepackage{orcidlink}
\def\confName{ICCV}
\def\confYear{2025}

\newcommand{\methname}{CoMet\xspace}

\newcolumntype{C}[1]{>{\centering}m{#1}}
\newcolumntype{L}[1]{>{}m{#1}}

\title{Towards Real Unsupervised Anomaly Detection Via Confident Meta-Learning}

\author{Muhammad Aqeel$^1$\orcidlink{0009-0000-5095-605X}\and Shakiba Sharifi$^1$\orcidlink{0009-0008-6309-635X} \and Marco Cristani$^{1,2}$\orcidlink{0000-0002-0523-6042} \and Francesco Setti$^{1,2}$\orcidlink{0000-0002-0015-5534} \\
$^1$ Dept. of Engineering for Innovation Medicine, University of Verona\\
Strada le Grazie 15, Verona, Italy\\
$^2$ Qualyco S.r.l., Strada le Grazie 15, Verona, Italy\\
{\tt\small Contact author: muhammad.aqeel@univr.it}
}
\begin{document}
\maketitle
\begin{abstract}

So-called unsupervised anomaly detection is better described as semi-supervised, as it assumes all training data are nominal. This assumption simplifies training but requires manual data curation, introducing bias and limiting adaptability. We propose \emph{Confident Meta-learning (\methname)}, a novel training strategy that enables deep anomaly detection models to learn from uncurated datasets where nominal and anomalous samples coexist, eliminating the need for explicit filtering.
Our approach integrates Soft Confident Learning, which assigns lower weights to low-confidence samples, and Meta-Learning, which stabilizes training by regularizing updates based on training-validation loss covariance. This prevents overfitting and enhances robustness to noisy data. \methname is model-agnostic and can be applied to any anomaly detection method trainable via gradient descent.
Experiments on MVTec-AD, VIADUCT, and KSDD2 with two state-of-the-art models 
demonstrate the effectiveness of our approach, consistently improving over the baseline methods, remaining insensitive to anomalies in the training set, and setting a new state-of-the-art across all datasets. Code is available at \url{https://github.com/aqeeelmirza/CoMet}
%

\end{abstract}

\section{Introduction}
\label{sec:intro}

\begin{figure}[t]
    \centering
    \includegraphics[width=1.0\linewidth]{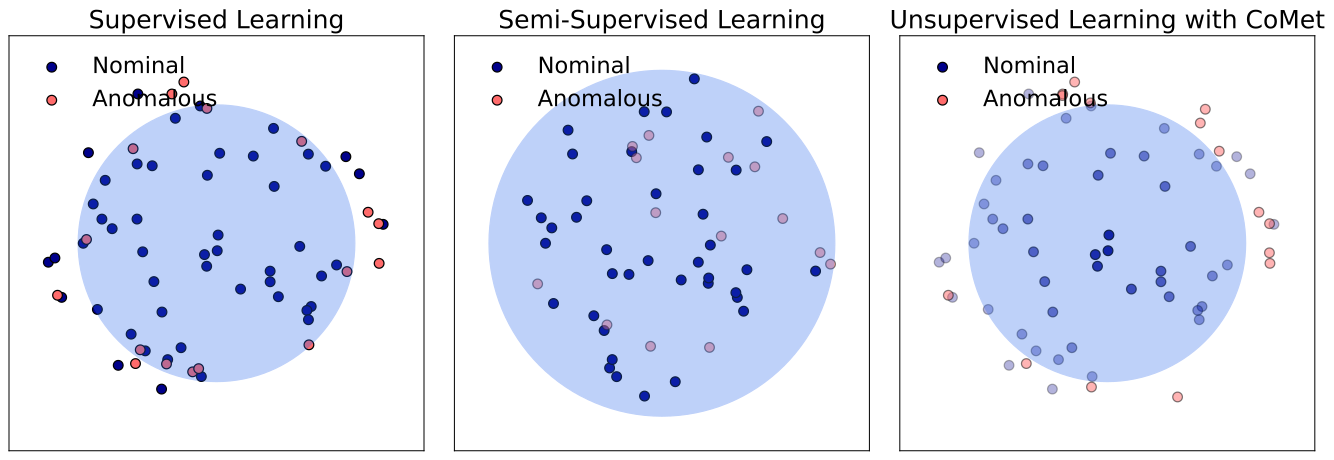}
    \caption{Impact of anomalies and near boundary samples in the training set. In Supervised Learning, positive and negative samples contribute to determining the boundary between the two data distributions. In Semi-Supervised Learning, often improperly called unsupervised, the lack of positive samples (\ie anomalies) lead to overestimate the covariance of the nominal samples' distribution, resulting in a higher number of misdetections at inference time (red points lying inside the blue circle). 
    Our Confident Meta-learning approach allows models to be learned properly in Unsupervised Learning by assigning low confidence weight to samples close to and beyond the decision boundary. Note that in Semi-Supervised Learning the positive samples (red dots) are transparent because they are not available at training time. In Unsupervised Learning instead, dots' transparency is proportional to the assigned confidence weights.}
    \label{fig:samples}
\end{figure}

In industrial manufacturing, real-time detection of defective products is critical to ensuring safety and quality standards, reducing waste, and optimizing production efficiency. The ability to detect defective products at an early stage prevents costly recalls and enhances customer satisfaction. Traditionally, quality control has relied on manual inspection by human operators. However, this process is expensive, produces inconsistent outputs, and is prone to errors due to fatigue and subjective judgment. Moreover, manual inspections are difficult to scale in high-throughput industrial settings. As a result, there is a strong demand for automated defect detection systems that can operate with high accuracy, reliability, and efficiency.  

Supervised learning approaches have shown strong performance in this task, but their effectiveness is highly dependent on the quantity and diversity of labeled defective and nominal samples available during training.
They require a substantial number of defective samples with pixel-level annotations. Since defects are inherently rare and often subtle, collecting and annotating such data is both costly and time-consuming. Moreover, many defect types exhibit high intra-class variability, requiring a sufficiently large and diverse dataset to ensure robust generalization across different defect categories.
To mitigate these challenges, recent research focused on unsupervised anomaly detection, where models are trained exclusively on nominal data (negative samples) and learn to identify deviations from the learned distribution. While this approach alleviates the need for labeled defective samples, it still requires an operator to carefully curate the training dataset to ensure that no anomalous samples are present. This manual filtering step introduces a significant limitation: it is time-consuming and susceptible to human error and bias. Even a small number of defective samples in the training set can lead the model to misclassify these defects as nominal, reducing its ability to detect true anomalies. Additionally, the assumption that all training samples are nominal does not hold in many real-world industrial applications, where undetected anomalies may be inadvertently included in the dataset. A graphical representation of this phenomenon is shown in Figure \ref{fig:samples}.

\begin{figure}[t]
    \centering
    \includegraphics[width=\linewidth]{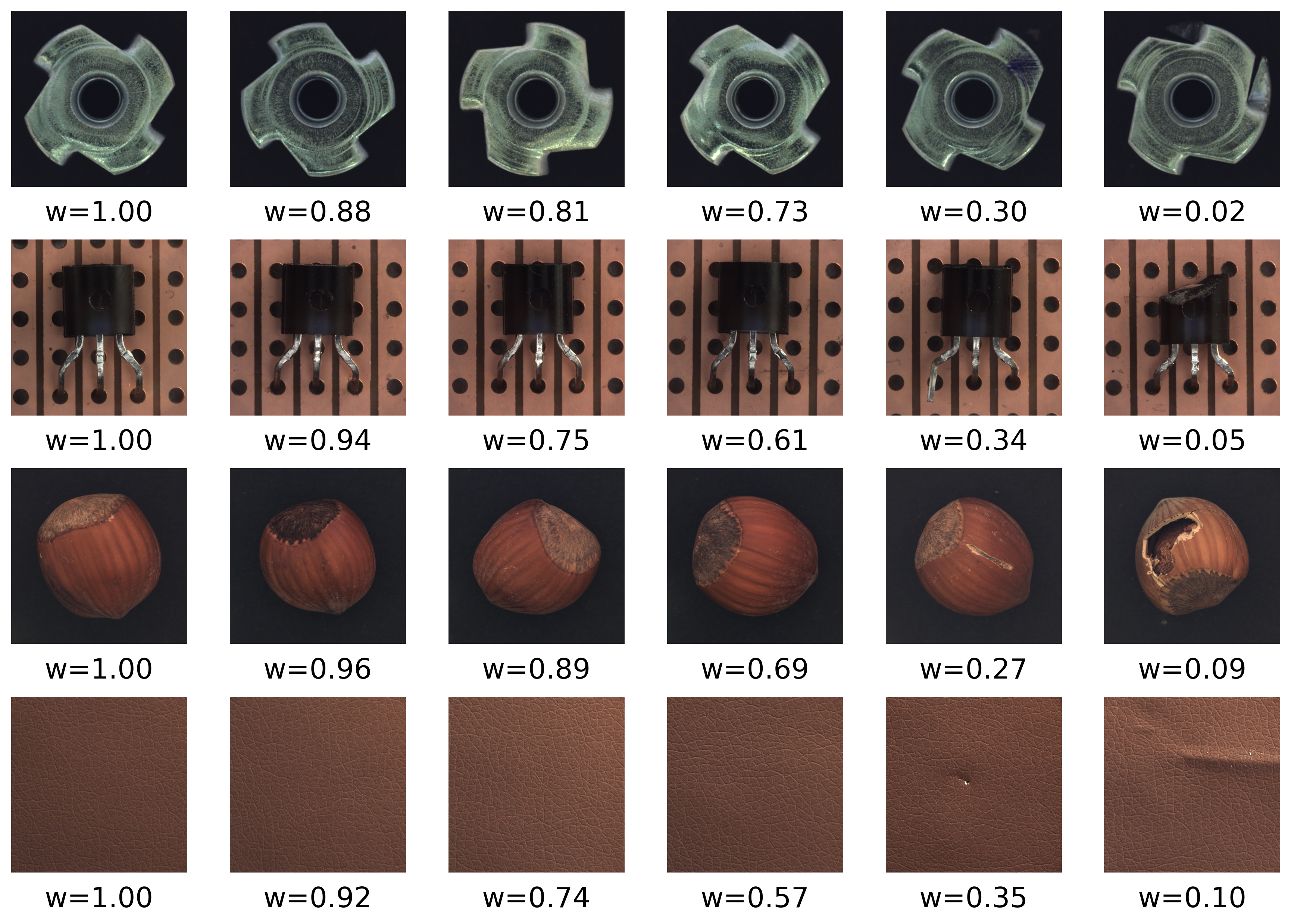}
    \caption{Training samples (nominal and anomalous) for some classes of the MVTec-AD dataset with the associated confidence weight $w$ estimated by \methname. Weights close to $1$ indicate prototypical samples, while lower weights suggest samples close to (or beyond) the decision boundary.}
    \label{fig:weight}
\end{figure}

In this paper, we argue that what is commonly referred to as \emph{unsupervised anomaly detection} is more accurately described as semi-supervised anomaly detection, as it implicitly assumes that all training samples are nominal, and we propose a novel training strategy that eliminates the need for manually filtering training data. Our method enables deep learning models to learn from raw, uncurated datasets where nominal and anomalous samples may coexist, without requiring explicit labels. By relaxing the assumption that all training data are nominal, our approach allows anomaly detection models to operate in a truly unsupervised manner. This not only reduces the burden of data annotation but also improves the model's adaptability to real-world conditions and enhances robustness against data distribution shifts.
%
We achieve this by leveraging a \emph{Soft Confident Learning} approach that dynamically estimates the reliability of training samples at each epoch based on the current model parameters. Samples exhibiting low confidence at a given epoch are assigned lower weights in subsequent gradient updates, reducing their influence on the learning process. Figure \ref{fig:weight} shows some confidence weights at the last training epoch. This strategy alone may introduce instability and increase the risk of overfitting. To mitigate these issues, we incorporate a \emph{Meta-Learning} framework that iteratively updates the model using partial subsets of training data while validating the learned representations on the remaining portion. The update step is regularized through a term inversely proportional to the covariance between training and validation losses, ensuring that large parameter updates are discouraged when the model exhibits overfitting tendencies. This combination of Soft Confident Learning and Meta-Learning enhances model robustness, allowing for more reliable learning in the presence of noisy or unfiltered training data.
Our \textbf{Confident Meta-learning} (\textbf{\methname}) approach can be used to extend to the truly unsupervised scenario any anomaly detection method that is trainable through gradient descent of a specific loss function, making it an extremely versatile approach that can adapt seamlessly to new state-of-the-art models.


Extensive experiments conducted on the popular benchmarks MVTec-AD~\cite{bergmann2019mvtec}, VIADUCT~\cite{lehr2024viaduct} and KSDD2~\cite{Bozic2021COMIND}, with two state-of-the-art anomaly detection models, namely DifferNet and SimpleNet, demonstrate the effectiveness of \methname in learning robust parametrizations. These parametrizations not only avoid overfitting to nominal samples but are also largely insensitive to the presence of anomalies in the training set.

The main contributions of our paper can be summarized as follows:
\begin{itemize}
    \setlength\itemsep{0em}
    \item We propose \methname, a novel training framework that allows anomaly detection models to learn more robust models by assigning low confidence scores to ambiguous samples near the decision boundary.
    \item Models trained with \methname achieve higher performance in anomaly detection by significantly reducing undetected anomalies (false negatives) at the cost of slightly increasing false positives.
    \item Extensive experiments on three public benchmarks demonstrate that \methname achieves state-of-the-art performances, effectively handling the presence of anomalous samples in the training set.  
\end{itemize}


\section{Related Work}
\label{sec:soa}

\begin{figure*}[t]
    \centering
    \includegraphics[width=\linewidth]{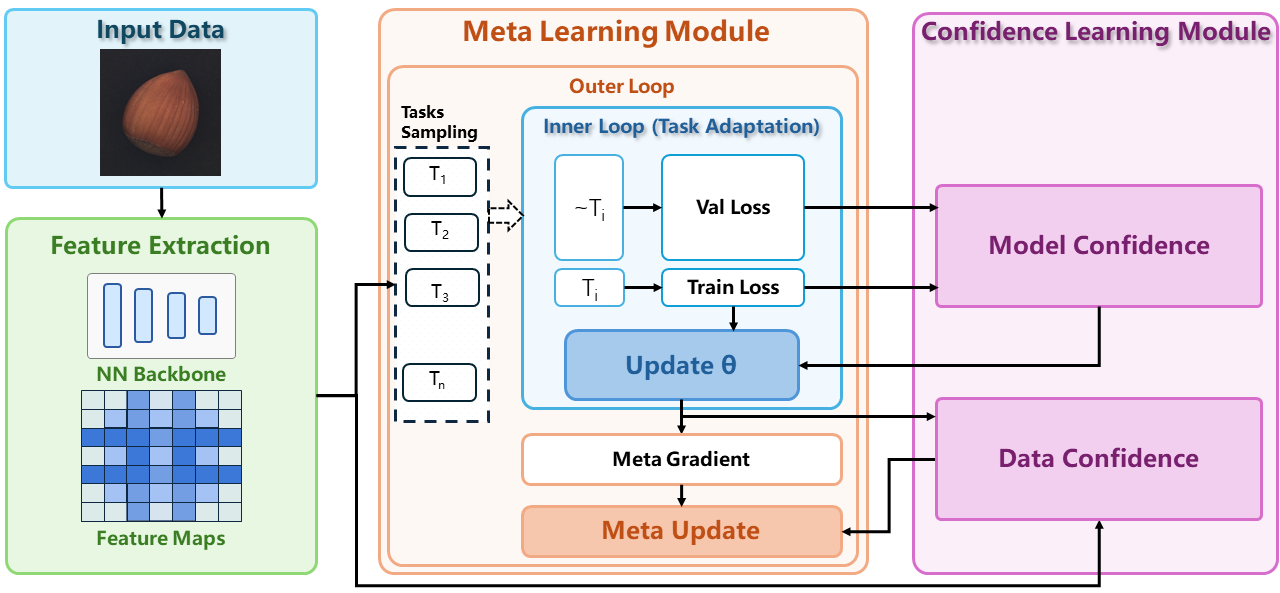}
    \caption{\methname pipeline. A pretrained feature extractor transforms input images to feature maps if required by the anomaly detection model of choice. Features are then divided into disjointed tasks. In the meta-learning inner loop one task is used for training and all the others for testing. Losses for training and validation are passed to the Soft Confident Learning module to compute the model confidence that is used for regularization of the weights update. Once all tasks have been processed, the SCL module uses current parametrization to compute the confidence of each sample data and use it to weight data in the meta update step.}
    \label{fig:pipeline}
\end{figure*}


Supervised learning methods have achieved remarkable results in industrial anomaly detection, mostly leveraging fully convolutional network architectures~\cite{baitieva2024segad,dong2019pganet}. However, collecting and annotating anomalous samples is a costly and labour-intensive operation that is also prone to errors. For this reason, recent research has focused on unsupervised learning techniques. Reconstruction-based methods learn a model that can compress and reconstruct the training data well, relying on the assumption that anomalous regions should not be properly reconstructed because not present in the training data. These approaches can work at the image level \cite{fuvcka2025transfusion,madan2023self,yao2024glad,zavrtanik2021draem} or at the feature level \cite{zavrtanik2022dsr}. Feature embedding-based methods use deep learning models pre-trained on large datasets to extract features from training samples. Different techniques are then applied to score anomalies by modeling data distributions \cite{RudWan2021,rudolph2022fully,yu2021fastflow,zhou2024msflow}, storing knowledge in memory banks \cite{defard2021padim,roth2022towards}, using teacher-student models \cite{batzner2024efficientad,rudolph2023asymmetric}, or training one-class discriminators \cite{li2021cutpaste,liu2023simplenet}.
Although commonly called unsupervised, a more accurate classification for these methods is semi-supervised, as they inherently assume that all training samples are nominal. While this assumption simplifies model training by eliminating the need for labeled anomalies, it still necessitates human intervention in selecting the training data, potentially introducing bias and limiting the model's robustness to unseen anomalies. To overcome this issue, some recent works used statistical methods to refine training data iteratively removing samples that are anomalous or close to the decision boundary \cite{aqeel2024selfirp,aqeel2025meta}. 
%
In this work, we address these limitations by employing meta-learning alongside soft confident learning to mitigate overfitting of problematic samples.

Soft Confident Learning (SCL) have been explored to enhance the reliability of supervised systems by identifying and correcting mislabeled samples in training datasets \cite{northcutt2021confident}. Rather than assuming all training labels are accurate, SCL aims to detect and re-weight or correct mislabeled examples. We extend SCL to the unsupervised scenario to assign low confidence scores to near- and beyond-boundary samples.

To improve model adaptability in dynamic environments, meta-learning has been extensively applied in scenarios requiring rapid model refinement, such as few-shot learning and domain adaptation \cite{finn2017model,hospedales2021meta}. Metric-based approaches, including prototypical networks \cite{snell2017prototypical} and matching networks \cite{vinyals2016matching}, learn an embedding space where similar samples are closer, facilitating fast adaptation with minimal data.
Optimization-based approaches learn an optimal initialization of the model that allows for quick adaptation to new tasks \cite{finn2017model}. These second strategy is more suited to the anomaly detection problem, where meta-learning enables models to adapt decision boundaries dynamically, allowing for effective handling of data distribution shifts over time \cite{Sung2018Learning}.



\section{Confident Meta-Learning pipeline}
\label{sec:method}

The proposed \methname approach leverages soft confident learning and meta-learning to perform anomaly detection within an unsupervised framework. 
The pipeline of the proposed approach is shown in Fig.~\ref{fig:pipeline}.

\subsection{Anomaly Detection backbone}

The \methname framework can be coupled with any anomaly detection model that can be trained end-to-end via gradient descent. In these terms, our approach is agnostic with respect to the anomaly detection backbone, and we will show experiments using two very different approaches based on normalizing flows and feature embeddings. 

Let $x_i \in \mathcal{D}_{train}$ be the set of training images consisting predominantly of nominal samples, with no explicit labels indicating anomalies.
During training, the AD model learns a function $\phi_{\theta}:\mathcal{D} \rightarrow \mathbb{R}$ that maps an input sample $x_i$ to an anomaly score $a_\theta(x_i)\in \mathbb{R}$, where higher scores indicate a higher likelihood of being anomalous. The parametrization $\theta$ is the one that minimized the method-specific loss function $\mathcal{L}_{AD}$: reconstruction-based methods typically minimize the Mean Squared Error (MSE), defined as $\mathcal{L}_{MSE} = \frac{1}{n} \sum (x_i - \hat{x}_i)^2$ with $\hat{x}_i$ the reconstructed input; density-based methods rely on Negative Log-Likelihood (NLL), given by $\mathcal{L}_{NLL} = -\sum \log (p(x_i | \theta))$ where $p(x | \theta)$ models the normal data distribution; and feature-based methods often incorporate variations of the Structural Similarity Index (SSIM), like $\mathcal{L}_{feat} = \sum \left( \phi_{\theta}(x_i)-c \right)^2$ where $c$ is the center of normal representations in feature space.
At testing time, an anomaly score $a_{\theta}(x_j)$ is computed for each test sample $x_j \in \mathcal{D}_{test}$, 
and an anomaly is flagged if $a_{\theta}(x_j) > \tau$, where $\tau$ is a predefined or adaptive threshold.

\subsection{Soft Confident Learning}

The goal of the soft confident learning module is to allow the model to rely more on those samples that are more prototypical for the nominal class and less on those samples that are anomalous or close to the boundary. To this aim, we quantify both model uncertainty and data uncertainty within an unsupervised learning framework, where we do not have access to labels. We will then use these uncertainties to estimate coefficients used in the training phase. 

\subsubsection{Quantifying Data Uncertainty}
To quantify data uncertainty, we adapt the concept of the \emph{confident joint} from the Confident Learning framework to our unsupervised setting. In this context, we consider the relationships between data points and their confidence scores assigned by the model. 

We notice that the anomaly score $a_{\theta}(x_i)$ reflects the confidence of a model parametrized by $\theta$ to predict $x_i$. Since we lack true labels, we use these confidence scores to assess the uncertainty associated with each data point.
We map these scores to weights to be used in a soft confident learning framework using a saturated inverse function of the confidence score as:
\begin{equation}
  w_i = \min(1,t/a_{\theta}(x_i))
\end{equation}
where $t$ is a threshold defined according to the Interquartile Range (IQR) statistical method to identify outliers~\cite{frery2023interquartile}. If we define $Q_1$ and $Q_3$ the first and third quartiles of the distribution of the confidence scores $a_{\theta}(x)$ respectively, the threshold $t$ is defined as:
\begin{equation} 
  t = Q_3 + \kappa (Q_3 - Q_1)
  \label{eq:thiqr}
\end{equation}
where $\kappa$ is a parameter that can be tuned to best suit the data characteristics.

We can now define the data-weighted loss function as:
\begin{equation} 
  \mathcal{L}_{data}(\theta) = \sum_{i=1}^{N} w_i \cdot \mathcal{L}_{AD}(x_i|\theta)  
\end{equation}
where $\mathcal{L}_{AD}(x_i|\theta)$ is the individual loss function of the anomaly detection model of choice. 

\subsubsection{Quantifying Model Uncertainty}

We quantify the model's uncertainty by calculating the determinant of the covariance matrix $\Sigma$ formed from the training and validation loss distributions. This approach aligns with statistical methods that interpret the determinant of a covariance matrix as a measure of the volume of spread in multivariate space, with larger values indicating greater variability or uncertainty \cite{kendall2017uncertainties}. Let $\mathcal{L}_{\text{train}}$ and $\mathcal{L}_{\text{val}}$ represent vectors of the model's training and validation losses, respectively. The covariance matrix $\Sigma$ is defined as:
\begin{equation}
  \Sigma = \begin{bmatrix} 
    \text{Cov}(\mathcal{L}_{\text{train}}, \mathcal{L}_{\text{train}}) & \text{Cov}(\mathcal{L}_{\text{train}}, \mathcal{L}_{\text{val}}) \\
    \text{Cov}(\mathcal{L}_{\text{val}}, \mathcal{L}_{\text{train}}) & \text{Cov}(\mathcal{L}_{\text{val}}, \mathcal{L}_{\text{val}})
  \end{bmatrix}.
\end{equation}

The determinant of $\Sigma$ serves as a scalar measure of the overall variability between training and validation losses: High values of $\det(\Sigma)$ indicate significant variability, suggesting high model uncertainty. Low values instead reflect consistent performance, indicating higher confidence in the learning process. 

To incorporate this measure into the training procedure, we introduce an adaptive regularization term $\lambda$ that adjusts dynamically:
\begin{equation}
  \lambda\left( \Sigma \right) = \lambda_0 \cdot \left( 1 + \gamma \cdot \det(\Sigma) \right)
  \label{eq:model}
\end{equation}
where $\lambda_0$ is the base regularization coefficient and $\gamma$ is a scaling hyperparameter controlling sensitivity to model uncertainty. This adaptive regularization allows the model to impose stronger regularization when uncertainty is high, promoting better generalization, and to relax regularization when the model is learning confidently.

\vspace{1em}
Combining both model and data uncertainty, our soft confident learning loss function becomes:
\begin{equation}
  \mathcal{L}_{SCL}(\theta) = \sum_{i=1}^{N} w_i \cdot \mathcal{L}_{AD}(x_i|\theta) + \lambda\left(\Sigma\right) \cdot \Vert \theta \Vert_2^2
  \label{eq:losscl}
\end{equation}

\subsection{Meta Learning}

\begin{figure}[t]
    \centering
    \includegraphics[width=.9\linewidth]{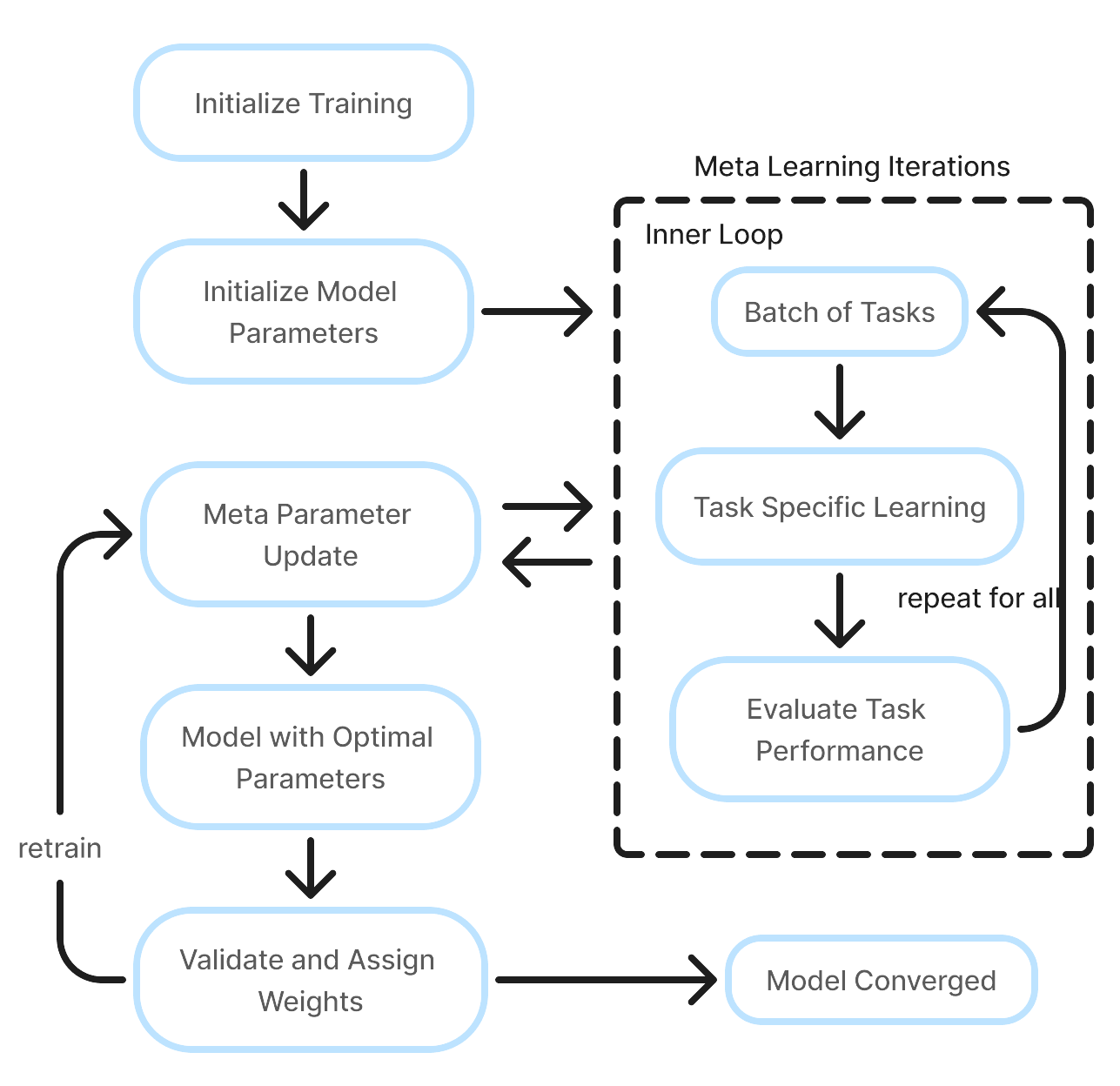}
    \caption{Meta-learning process with outer and inner loops. In the inner loop, each task batch $T_1$ to $T_n$ updates task-specific parameters by optimizing $\theta$. After completing inner loop updates, the outer loop aggregates these adjustments to refine the global $\theta$, enhancing the model's ability to generalize and adapt quickly to new tasks.}
    \label{fig:meta}
\end{figure}

To enhance the model's ability to efficiently adapt to new data with sample-specific confidence, we incorporate Model-Agnostic Meta-Learning (MAML) \cite{finn2017model} into our unsupervised framework. Unlike standard regularization that treats parameters uniformly, MAML enables covariance-based uncertainty quantification and identifies which parameter updates generalize well across tasks. We randomly divide training samples into $n$ disjoint tasks, where each iteration uses task $T_i$ for training and the remaining tasks for validation, creating the structured splits necessary for our adaptive regularization term $\lambda(\Sigma)$.

The MAML algorithm, shown in Figure~\ref{fig:meta}, consists of two optimization loops. In the inner loop, we update the model parameters to fit a specific task using gradient descent steps with learning rate $\alpha=10^{-4}$:
\begin{equation}
  \theta' = \theta - \alpha \nabla_{\theta} \mathcal{L}_{\text{train}}(\theta),
\end{equation}
where $\mathcal{L}_{\text{train}}(\theta)$ is the training loss on the current task computed as in \eqref{eq:losscl}.

In the outer loop, we update the original model parameters $\theta$ using validation performance across the remaining tasks with meta-learning rate $\beta=2\times10^{-4}$:
\begin{equation}
  \theta \leftarrow \theta - \beta \nabla_{\theta} \mathcal{L}_{\text{meta}}(\theta'),
\end{equation}
This two-level optimization enables the model to learn which parameter updates generalize well, effectively learning sample-specific weights while balancing confidence-weighted updates with generalization performance.

We integrate the reweighted loss function from Soft Confident Learning into the meta-objective:
\begin{equation}
\mathcal{L}_{\text{meta}}(\theta') = \sum_{i=1}^{N} w_i \cdot \mathcal{L}_{AD}(x_i|\theta') + \lambda\left(\Sigma'\right) \cdot \Vert \theta' \Vert_2^2,
\end{equation}
where $\theta'$ are the adapted parameters from the inner loop, $\mathcal{L}_{AD}(x_i|\theta')$ is the individual loss for data point $x_i$ using parameters $\theta'$, $w_i$ are the weights based on confidence scores calculated using $\theta'$, and $\lambda(\Sigma')$ is the adaptive regularization term as previously defined in \eqref{eq:model}.

\section{Experiments}
\label{sec:exp}



We present experiments integrating \methname with two distinct approaches, enhancing their performance. Using the standard protocol with only nominal training data, our method outperforms baselines by isolating boundary-close nominal samples. We also demonstrate robustness to mislabelled anomalies in training, and an ablation study confirms the impact of our proposed modules.

\subsection{Datasets}
We extensively validated our proposed approach on three challenging public benchmarks. 
We report results on the \textbf{MVTec-AD} dataset~\cite{bergmann2019mvtec}, which consists of 5,354 high-resolution color images spanning over 10 object and 5 texture categories, each containing multiple defect types. While this dataset remains the most widely used benchmark in the field, it is now considered largely solved, with several methods achieving AUROC scores surpassing 99\%. 
A more challenging scenario is introduced by the recently released \textbf{VIADUCT} dataset~\cite{lehr2024viaduct}, a benchmark consisting of 10,986 high-resolution images across 49 categories from 11 distinct sectors. Each object is captured from five different perspectives, offering a comprehensive view of the dataset's variability. 
Finally, the \textbf{KSDD2} dataset~\cite{Bozic2021COMIND} is a surface anomaly detection dataset containing 2,085 anomaly-free images and 246 anomalous images. The dataset is particularly challenging due to the presence of near-in-distribution surface anomalies, making it difficult to distinguish true anomalies from typical surface variations.

\subsection{Evaluation Metrics}
We evaluate image-level anomaly detection performance using the standard Area Under the Receiver Operating Characteristic Curve (AUROC), denoted as I-AUROC, based on the anomaly detection scores $a_{\theta}(x)$, equations \eqref{eq:scoreNF} and \eqref{eq:scoreSN}. We also report pixel-level AUROC (P-AUROC).

\subsection{Implementation Details}
We implemented our model using the PyTorch framework and trained it on an NVIDIA RTX 4090 GPU for efficient training and inference. Input images were resized to $256 \times 256$ pixels with optional rotation augmentations. 
Regularization was achieved through weight decay to prevent overfitting. Training was conducted over 180 epochs with a batch size of 192 and a learning rate of $2 \times 10^{-4}$. This setup enables our model to effectively learn and generalize complex data distributions, achieving robust performance in density estimation and generative tasks.

\begin{table*}[ht]
\centering
\caption{Comparison results on the MVTec-AD dataset of anomaly detection performance (I-AUROC/{\color{gray}P-AUROC} in \%).}
\label{tab:mvtec} \vspace{-.5em}
\resizebox{\textwidth}{!}{%
\begin{tabular}{L{2.7cm}*{4}{C{2.1cm}}|cc|cc}
\toprule
\textbf{Method} & \textbf{PaDiM}\cite{defard2021padim} & \textbf{DRÆM}\cite{zavrtanik2021draem} & \textbf{CSFlow}\cite{rudolph2022fully} & \textbf{PatchCore}\cite{roth2022towards} & \textbf{DifferNet}\cite{RudWan2021} & \textbf{\methname-NF} & \textbf{SimpleNet}\cite{liu2023simplenet} & \textbf{\methname-SN} \\ 
\midrule
\textbf{Carpet}      & 99.8/{\color{gray}99.1} & 97.0/{\color{gray}95.5} & \textbf{100}/- & 98.7/{\color{gray}99.0} & 92.9/{\color{gray}90.7} & 98.2/{\color{gray}96.5} & 99.7/{\color{gray}98.2} & 99.7/{\color{gray}98.3} \\
\textbf{Grid}        & 96.7/{\color{gray}97.3} & 99.9/{\color{gray}97.7} & 99.0/- & 98.2/{\color{gray}98.7} & 84.0/{\color{gray}89.9} & 97.5/{\color{gray}96.6} & 99.7/{\color{gray}98.8} & \textbf{100}/{\color{gray}99.0} \\
\textbf{Leather}     & \textbf{100}/{\color{gray}99.2} & \textbf{100}/{\color{gray}98.6} & \textbf{100}/- & \textbf{100}/{\color{gray}99.3} & 97.1/{\color{gray}98.1} & \textbf{100}/{\color{gray}98.2} & \textbf{100}/{\color{gray}99.2} & \textbf{100}/{\color{gray}99.8} \\
\textbf{Tile}        & 98.1/{\color{gray}94.1} & 99.6/{\color{gray}99.2} & \textbf{100}/- & 98.7/{\color{gray}95.6} & 99.4/{\color{gray}96.3} & \textbf{100}/{\color{gray}99.2} & 99.8/{\color{gray}97.0} & \textbf{100}/{\color{gray}94.1} \\
\textbf{Wood}        & 99.2/{\color{gray}94.9} & 99.1/{\color{gray}96.4} & \textbf{100}/- & 99.2/{\color{gray}95.0} & 99.8/{\color{gray}98.2} & \textbf{100}/{\color{gray}99.8} & \textbf{100}/{\color{gray}94.5} & \textbf{100}/{\color{gray}94.4} \\
\midrule
\textbf{Bottle}      & 99.1/{\color{gray}98.3} & 99.2/{\color{gray}99.1} & 99.8/- & \textbf{100}/{\color{gray}98.6} & 99.0/{\color{gray}98.3} & \textbf{100}/{\color{gray}99.8} & \textbf{100}/{\color{gray}96.9} & \textbf{100}/{\color{gray}98.6} \\
\textbf{Cable}       & 97.1/{\color{gray}96.7} & 91.8/{\color{gray}94.7} & 99.1/- & 99.5/{\color{gray}98.4} & 95.9/{\color{gray}94.6} & 98.9/{\color{gray}97.6} & 99.9/{\color{gray}97.6} & \textbf{100}/{\color{gray}98.1} \\
\textbf{Capsule}     & 87.5/{\color{gray}98.5} & \textbf{98.5}/{\color{gray}94.3} & 97.1/- & 98.1/{\color{gray}98.8} & 86.9/{\color{gray}96.7} & 98.1/{\color{gray}96.4} & 97.7/{\color{gray}98.9} & 98.1/{\color{gray}99.6} \\
\textbf{Hazelnut}    & 99.4/{\color{gray}98.2} & \textbf{100}/{\color{gray}99.7} & 99.6/- & 99.9/{\color{gray}98.7} & 99.3/{\color{gray}98.4} & \textbf{100}/{\color{gray}99.6} & \textbf{100}/{\color{gray}97.9} & \textbf{100}/{\color{gray}98.3} \\
\textbf{Metal Nut}   & 96.2/{\color{gray}97.2} & 98.7/{\color{gray}99.5} & 99.1/- & \textbf{100}/{\color{gray}98.4} & 96.1/{\color{gray}97.9} & 99.7/{\color{gray}98.9} & \textbf{100}/{\color{gray}98.8} & \textbf{100}/{\color{gray}99.4} \\
\textbf{Pill}        & 90.1/{\color{gray}95.7} & 98.9/{\color{gray}97.6} & 98.6/- & 96.6/{\color{gray}97.4} & 88.8/{\color{gray}95.3} & 98.6/{\color{gray}96.8} & \textbf{99.0}/{\color{gray}95.1} & \textbf{99.0}/{\color{gray}98.9} \\
\textbf{Screw}       & 97.5/{\color{gray}98.5} & 93.9/{\color{gray}97.6} & 97.6/- & 98.1/{\color{gray}99.4} & 96.3/{\color{gray}96.7} & 98.3/{\color{gray}96.6} & 98.2/{\color{gray}99.3} & \textbf{98.7}/{\color{gray}99.8} \\
\textbf{Toothbrush}  & \textbf{100}/{\color{gray}98.8} & \textbf{100}/{\color{gray}98.1} & 91.9/- & \textbf{100}/{\color{gray}98.7} & 98.6/{\color{gray}99.0} & \textbf{100}/{\color{gray}100} & 99.7/{\color{gray}98.5} & \textbf{100}/{\color{gray}99.1} \\
\textbf{Transistor}  & 94.4/{\color{gray}97.5} & 93.1/{\color{gray}90.9} & 99.3/- & \textbf{100}/{\color{gray}96.3} & 91.1/{\color{gray}93.2} & 99.1/{\color{gray}98.6} & \textbf{100}/{\color{gray}97.6} & \textbf{100}/{\color{gray}97.5} \\
\textbf{Zipper}      & 98.6/{\color{gray}98.5} & \textbf{100}/{\color{gray}98.8} & 99.7/- & 99.4/{\color{gray}98.8} & 95.1/{\color{gray}96.4} & 99.6/{\color{gray}96.8} & 99.9/{\color{gray}98.9} & \textbf{100}/{\color{gray}99.4} \\
\midrule
\textit{Average}     & 95.8/{\color{gray}97.5} & 98.0/{\color{gray}97.3} & 98.7/- & 99.1/{\color{gray}98.1} & 94.9/{\color{gray}96.0} & \textbf{99.2}/{\color{gray}98.1} & 99.6/{\color{gray}98.1} & \textbf{99.7}/{\color{gray}98.3} \\ 
\bottomrule
\end{tabular}%
}

\vspace{1em}
\caption{Results of anomaly detection on VIADUCT dataset (I-AUROC/{\color{gray}P-AUROC} in \%).}
\label{tab:viaduct} \vspace{-.5em}
\resizebox{\textwidth}{!}{%
\begin{tabular}{L{2.7cm}*{4}{C{2.1cm}}|cc|cc}
\toprule
\textbf{Method} & \textbf{EfficientAD}\cite{batzner2024efficientad} & \textbf{DRÆM}\cite{zavrtanik2021draem} & \textbf{MSFlow}\cite{zhou2024msflow} & \textbf{PatchCore}\cite{roth2022towards} & \textbf{DifferNet}\cite{RudWan2021} & \textbf{\methname-NF} & \textbf{SimpleNet}\cite{liu2023simplenet} & \textbf{\methname-SN} \\ 
\midrule
\textbf{Shredded CR.} & 84.1/- & 51.4/- & 94.1/- & 89.1/- & 80.9/{\color{gray}92.6} & 85.2/{\color{gray}88.6} & 95.3/{\color{gray}96.4} & \textbf{98.5}/{\color{gray}99.3}\\
\textbf{Encoder} & 51.2/- & 63.0/- & 51.9/- & 70.9/- & 51.5/{\color{gray}53.6} & 64.3/{\color{gray}61.2} & 80.2/{\color{gray}79.9} & \textbf{84.7}/{\color{gray}97.3}\\
\textbf{Raspberry} & 94.7/- & 90.0/- & 92.8/- & \textbf{98.9}/- & 91.0/{\color{gray}92.1} & 93.6/{\color{gray}92.2} & 97.6/{\color{gray}99.0} & 98.6/{\color{gray}99.1}\\
\textbf{Device Box} & 87.5/- & 50.8/- & 79.6/- & 75.3/- & 55.8/{\color{gray}60.7} & 72.7/{\color{gray}68.4} & 89.0/{\color{gray}97.0} & \textbf{95.0}/{\color{gray}98.2}\\
\textbf{L-Fitting} & 88.8/- & 55.7/- & 69.6/- & 80.6/- & 80.8/{\color{gray}85.4} & 82.4/{\color{gray}80.2} & 85.4/{\color{gray}96.9} & \textbf{95.4}/{\color{gray}99.3}\\
\textbf{Threaded Fitting} & 59.6/- & 60.5/- & 58.6/- & 55.0/- & 56.5/{\color{gray}64.7} & \textbf{64.9}/{\color{gray}64.7} & 58.9/{\color{gray}70.3} & 61.3/{\color{gray}79.8}\\
\textbf{Redon needle} & 98.1/- & 66.8/- & \textbf{100}/- & 97.2/- & 93.2/{\color{gray}92.7} & \textbf{100}/{\color{gray}95.8} & 99.9/{\color{gray}99.9} & \textbf{100}/{\color{gray}99.1}\\
\textbf{Aluminium Plate} & 89.4/- & 56.7/- & \textbf{96.2}/- & 86.6/- & 85.9/{\color{gray}98.8} & 92.3/{\color{gray}\textbf{93.5}} & 92.7/{\color{gray}94.5} & 96.0/{\color{gray}98.9}\\
\textbf{PaperClip} & 97.2/- & 69.5/- & 99.2/- & 99.8/- & 90.8/{\color{gray}92.6} & 97.9/{\color{gray}94.2} & 99.5/{\color{gray}95.1} & \textbf{100}/{\color{gray}97.6}\\
\textbf{Air Muffler Large} & 89.1/- & 55.4/- & 94.7/- & \textbf{99.8}/- & 89.9/{\color{gray}91.6} & 94.5/{\color{gray}91.6} & 99.0/{\color{gray}99.3} & 99.7/{\color{gray}99.4}\\
\textbf{Saw Blade} & 69.2/- & 58.2/- & 70.2/- & 61.0/- & 61.8/{\color{gray}64.7} & \textbf{70.3}/{\color{gray}66.5} & 60.8/{\color{gray}89.7} & 63.9/{\color{gray}98.1}\\
\midrule
\textit{Average} & 82.6/- & 61.6/- & 82.4/- & 83.1/- & 76.2/{\color{gray}80.9} & \textbf{83.5}/{\color{gray}81.5} & 87.1/{\color{gray}92.5} & \textbf{90.3}/{\color{gray}97.0}\\ 
\bottomrule
\end{tabular}%
}

\vspace{1em}
\caption{Comparison results on the KSDD2 dataset of anomaly detection performance (I-AUROC/{\color{gray}P-AUROC} in \%).}
\label{tab:ksdd2} \vspace{-.5em}
\resizebox{\textwidth}{!}{%
\begin{tabular}{L{2.7cm}*{4}{C{2.1cm}}|cc|cc}
\toprule
\textbf{Method} & \textbf{DRAEM}\cite{zavrtanik2021draem} & \textbf{MAD}\cite{madan2023self} & \textbf{DSR}\cite{zavrtanik2022dsr} & \textbf{MLD-IR}\cite{aqeel2025meta} & \textbf{DifferNet}\cite{RudWan2021} & \textbf{\methname-NF} & \textbf{SimpleNet}\cite{liu2023simplenet} & \textbf{\methname-SN} \\ 
\midrule
\textit{Average}     & 77.8/{\color{gray}--} & 79.3/{\color{gray}--} & 87.2/{\color{gray}--} & 94.3/{\color{gray}92.6} & 91.5/{\color{gray}92.1} & \textbf{94.9}/{\color{gray}93.4} & 91.7/{\color{gray}93.0} & \textbf{92.2}/{\color{gray}93.4} \\ 
\bottomrule
\end{tabular}%
}
\end{table*}

\subsection{\methname with Normalizing Flows}
\label{sec:nf}

Normalizing flows (NF) are known to be effective in mapping input data $x\in X$ sampled from a complex distribution $p(x)$ to a latent space $z$ with a simpler, \eg gaussian, distribution $p(z)$. This is achieved by a series of transformations parametrized by $\theta$. The mapping of data distributions is then defined by:
\begin{equation}
  p_{\theta}(x) = p_{\theta}(z) \left| \det \frac{\partial z}{\partial u} \right|
\end{equation}

In our work we incorporate the DifferNet model proposed by \cite{RudWan2021}: we use a pre-trained AlexNet to map training images $x_i\in X$ to a multi-scale feature space $u_i\in U$, capturing both fine and coarse details. 
Each input distribution is split into two parts, \( u_1 \) and \( u_2 \), which interact with each other through alternating translation ($\tau$) and scale ($\sigma$) functions. 
During training, the normalizing flow model is optimized to find the parametrization $\theta$ that minimizes the probability distribution of the inputs' negative log-likelihood of nominal samples. 
The loss function is:
\begin{equation}
  \mathcal{L}_{NF}(u|\theta) = \frac{\|z\|^2}{2} - \log \left| \det \frac{\partial z}{\partial u} \right|
  \label{eq:lossNF}
\end{equation}
where the first term encourages features to map close to $z$=0 in the latent space, and the second term, involving the Jacobian's log determinant, penalizes trivial scaling solutions, thereby promoting meaningful transformation. By applying multiple transformations of each input during training, the model learns a robust mapping that generalizes well.

At inference time, the NF model evaluates the likelihood of image features. Features with low likelihood are flagged as anomalies. 
The scoring function is defined as:
\begin{equation}
    a_{\theta}(x_i) = \mathbb{E}_{S_i} \left[ - \log p \left( f_{\theta} \left( f_{\phi} \left( S_i(x_i) \right) \right) \right) \right]
    \label{eq:scoreNF}
\end{equation}
where $S_i$ is a transformation applied to the input $x_i$ (\eg, rotations, translations, flips), allowing for a more stable anomaly score across variations. $f_{\phi}$ is the pre-trained feature extractor that maps the input images to a multi-scale feature space, and $f_{\theta}$ is the Normalizing Flow model. 
The anomaly score $a_{\theta}(x_i)$ is finally the expected value of the negative log-likelihood over multiple transformations $S_i$, ensuring robustness to small changes in the input.

\subsection{\methname with SimpleNet}
\label{sec:simplenet}

SimpleNet \cite{liu2023simplenet} is a lightweight anomaly detection method that operates directly in the feature space. We use the original formulation of this method that employs a wideResNet-50 architecture pre-trained on Imagenet to map input images $x_i$ to local features $o_i=F_*(x_i)$, where the * symbol indicates that the parametrization of the feature extractor is frozen.
A shallow neural network $G_{\theta_1}$, usually a single fully-connected layer, adapts the extracted features to a lower-dimensional space, also reducing domain bias and producing task-specific features $q_i$:
\begin{equation}
    q_i = G_{\theta_1}(o_i) = G_{\theta_1}(F_*(x_i))
\end{equation}

At training time, SimpleNet generates synthetic anomalies by perturbing normal features with Gaussian noise such as:
\begin{equation}
    q_i^- = q_i + \epsilon \; , \;\; \epsilon \sim\mathcal{N}(0,\sigma^2)
\end{equation}
where $\sigma$ controls the perturbation intensity.

Finally, a binary discriminator $D_{\theta_2}$ is trained to act as a normality scorer, estimating the normality of each pixel in the image. This is a 2-layer MLP structure that learns to output positive scores for normal features, and negative scores for generated anomalous features.
The loss function is based on the truncated $L_1$ norm as follows:
\begin{equation}
    \mathcal{L}(x|\theta) = \max(0,th-D_{\theta_2}(q_i)) + \max(0,-th+D_{\theta_2}(q_i)+\epsilon)
    \label{eq:lossSN}
\end{equation}
with $th$ a saturation threshold set to 0.5 by default, and $\theta=\{\theta_1,\theta_2\}$ is the complete parametrization of the model accounting for the feature adapter and the discriminator.

During inference, the anomaly feature generator is removed, and the discriminator directly outputs the anomaly score:
\begin{equation}
    a_{\theta}(x_i) = D_{\theta_2}(G_{\theta_1}(F_*(x_i))) .
    \label{eq:scoreSN}
\end{equation}

\subsection{Results}
\label{result}

Quantitative results on each class of MVTecAD, VIADUCT, and KSDD2 datasets at image- (I-AUROC) and pixel-level (P-AUROC) are reported in Tables~\ref{tab:mvtec}, \ref{tab:viaduct}, and \ref{tab:ksdd2} respectively. We report the results of the four best-performing state-of-the-art methods, and a direct comparison with DifferNet and SimpleNet with our \methname training procedure.

Our method demonstrates competitive performance across all the benchmarks. On MVTec-AD dataset (Table~\ref{tab:mvtec}), we achieve the best performance of 99.7\% using \methname-SN, the best performance on this dataset. 
The improvement on the original SimpleNet model is just 0.1\%, yet a great achievement considering how close we are to the perfect performance.
Moreover, our method consistently outperforms the baseline on all 15 classes, with a perfect score on 11 of them. \methname-SN also sets the new state-of-the-art in terms of P-AUROC with 98.3\%. As for the normalizing flows, \methname-NF is performing very well with 99.2\% I-AUROC on average and an improvement over the DifferNet of 4.3\%. The performance is largely due to poor performance on \textsc{Grid}, \textsc{Capsule}, and \textsc{Pill}, which are also problematic for the DifferNet method, suggesting that the challenge may lie in fitting the reconstruction using normalizing flows.

On the VIADUCT dataset (Table~\ref{tab:viaduct}), \methname-SN establishes a new state-of-the-art performance, outperforming all methods in 6 out of 11 classes. Among the remaining 5 classes, in 2 of them the best performance is achieved by \methname-NF, witnessing the effectiveness of our proposed training strategy; for the last 3 classes, our \methname-SN is always the runner-up with an average gap of only 0.2\%. With an overall average I-AUROC of 83.5\% for \methname-NF (7.3\% better than the DifferNet) and 90.3\% for \methname-SN (3.2\% better than SimpleNet) across 11 different sectors, our method demonstrates strong consistency in handling diverse anomalies and adaptability to challenging scenarios.

Finally, on the KSDD2 dataset (Table~\ref{tab:ksdd2}), our approach sets the new state-of-the-art I-AUROC with 94.9\% using normalizing flows model (\methname-NF), while the performance improvement of \methname-SN is more limited in this case. This is possibly due to the fact that DifferNet is particularly suited for this kind of texture, as proved by the high results achieved by the baseline method. Both variants are comparable to the state of the art. 

Across all datasets, our method consistently establishes a new state-of-the-art in both image-level and pixel-level AUROC scores, demonstrating robust performance across diverse object types and anomaly complexities. Additionally, it significantly outperforms baseline methods across both metrics, achieving substantial improvements in accuracy and reliability.
We attribute this strong performance to our framework’s ability to prioritize real prototypical samples while reducing the influence of nominal samples near the decision boundary. This leads to fewer false negatives (undetected anomalies) at the cost of a slight increase in false positives, \ie nominal samples incorrectly flagged as anomalous. This is supported by the results in Table \ref{tab:precrec}, which show \methname significantly improves recall while maintaining a controlled decrease in precision.

\begin{table}[t]
    \centering
    \caption{Average Precision and Recall for baseline and \methname models on MVTec AD, VIADUCT and KSDD2 datasets.} \vspace{-0.5em}
    \label{tab:precrec}
    \resizebox{\linewidth}{!}{
    \begin{tabular}{lccc|ccc}
        \hline
        \textbf{Dataset} & \multicolumn{3}{c|}{\textbf{DifferNet} \cite{RudWan2021}} & \multicolumn{3}{c}{\textbf{\methname-NF}}  \\
        & \textbf{Precision} & \textbf{Recall} & \textbf{F$_1$-score} & \textbf{Precision} & \textbf{Recall} & \textbf{F$_1$-score} \\
        \hline
        MVTec AD & 95.6 & 76.4 & 84.9 & 92.5 & 93.4 & 92.9 \\
        VIADUCT & 79.1 & 70.3 & 74.4 & 77.1 & 90.8 & 83.4 \\
        KSDD2 & 90.9 & 87.5 & 89.2 & 87.4 & 94.3 & 90.7 \\
        \hline
        \hline
        \textbf{Dataset} & \multicolumn{3}{c|}{\textbf{SimpleNet} \cite{liu2023simplenet}} & \multicolumn{3}{c}{\textbf{\methname-SN}}  \\
        & \textbf{Precision} & \textbf{Recall} & \textbf{F$_1$-score} & \textbf{Precision} & \textbf{Recall} & \textbf{F$_1$-score} \\
        \hline
        MVTec AD & 98.1 & 98.9 & 98.5 & 97.8 & 99.8 & 98.8 \\ 
        VIADUCT & 84.1 & 93.6 & 88.6 & 83.4 & 97.5 & 89.9 \\ 
        KSDD2 & 96.2 & 70.0 & 81.0 & 96.0 & 70.3 & 81.2 \\ 
        \hline
    \end{tabular}
    }
\end{table}

\subsection{Ablation Study}
\label{sec:ablation}

\begin{table}[t]
\centering
\caption{Average I-AUROC on MVTec-AD dataset for different component removals in the ablation study.} \vspace{-0.5em}
\label{tab:ablation}
\resizebox{\linewidth}{!}{%
\begin{tabular}{lcc}
\hline
\textbf{Configuration} & \textbf{\methname-NF} & \textbf{\methname-SN} \\
\hline
\textbf{\methname w/o SCL and ML (baseline)} & 94.9 & 99.4 \\
\textbf{\methname w/o ML} & 96.8 & 99.4\\
\textbf{\methname w/o SCL on Data \& Model} & 97.2 & 99.5\\
\textbf{\methname w/o SCL on Data} & 97.9 & 99.5 \\
\textbf{\methname (full)} & 99.2 & 99.7 \\
\hline
\end{tabular}%
}
\end{table}

To assess the contribution of individual components in our framework, we conducted an ablation study on the MVTec-AD dataset, systematically removing Meta-Learning (ML) and Soft Confident Learning (SCL) from our model. The results, presented in Table \ref{tab:ablation}, highlight the significant impact of these components on overall performance.
Removing the Meta-Learning component (\textbf{w/o ML}) results in a significant decrease in I-AUROC of about 2.5\% on NF model and 0.3\% for SN model.
Indeed, without meta-learning, the training and validation sets coincide at each step, making the model susceptible to overfitting when computing the confidence weights.
Nevertheless, this model achieves a substantial improvement over the baseline for NF, demonstrating the effectiveness of the Soft Confident Learning module on its own.
Removing Soft Confident Learning on Data (\textbf{w/o SCL on Data}) lowers I-AUROC, highlighting its role in handling ambiguous samples. Omitting both Soft Confident Learning on Data and Model (\textbf{w/o SCL on Data \& Model}) further reduces performance, demonstrating the added benefit of integrating SCL at both levels.
%

\subsection{Noise Robustness}

\begin{figure}[t]
    \centering
    \begin{subfigure}{0.49\linewidth}
        \centering
        \includegraphics[width=\linewidth]{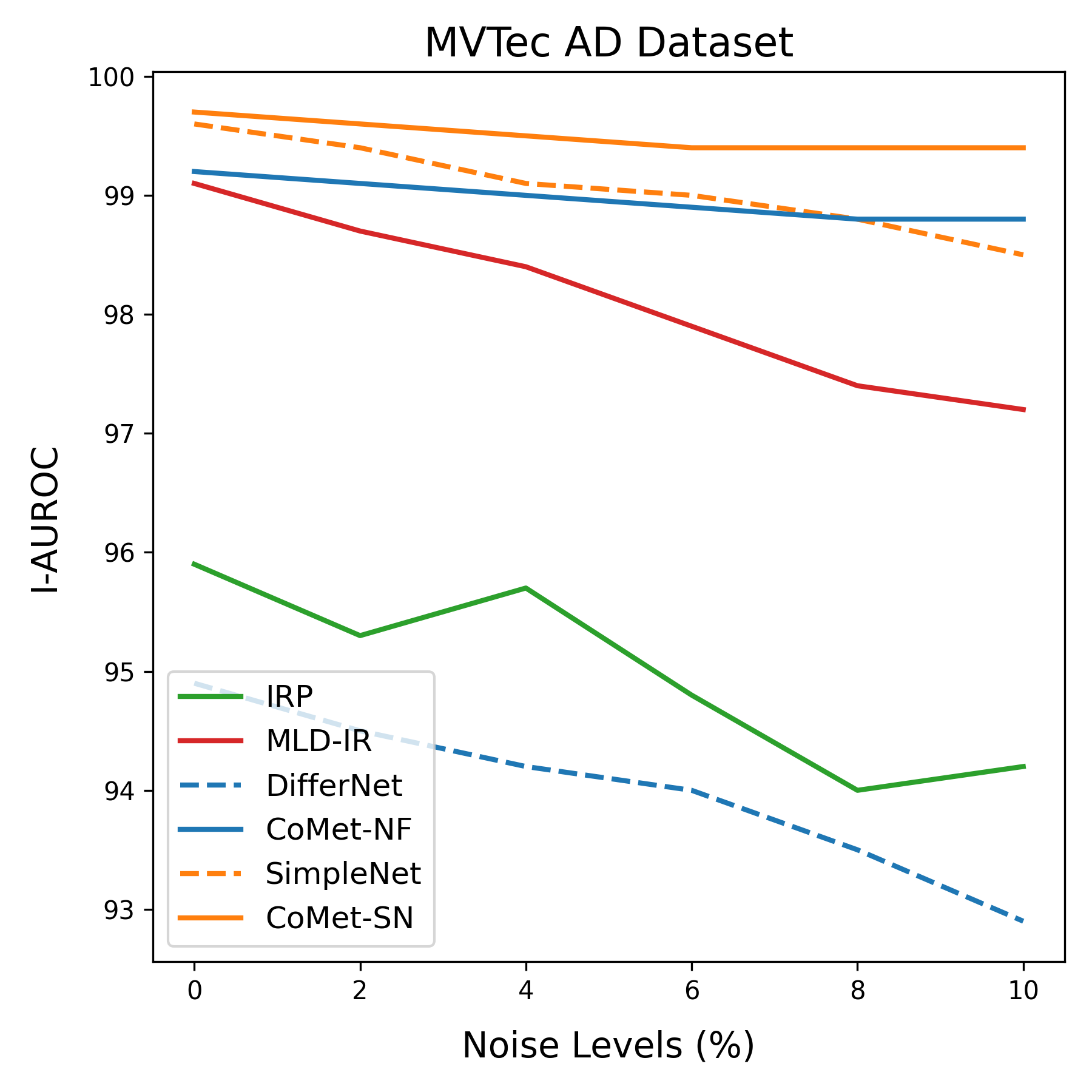}
        \caption{}
        \label{fig:image1}
    \end{subfigure}
    \hfill
    \begin{subfigure}{0.49\linewidth}
        \centering
        \includegraphics[width=\linewidth]{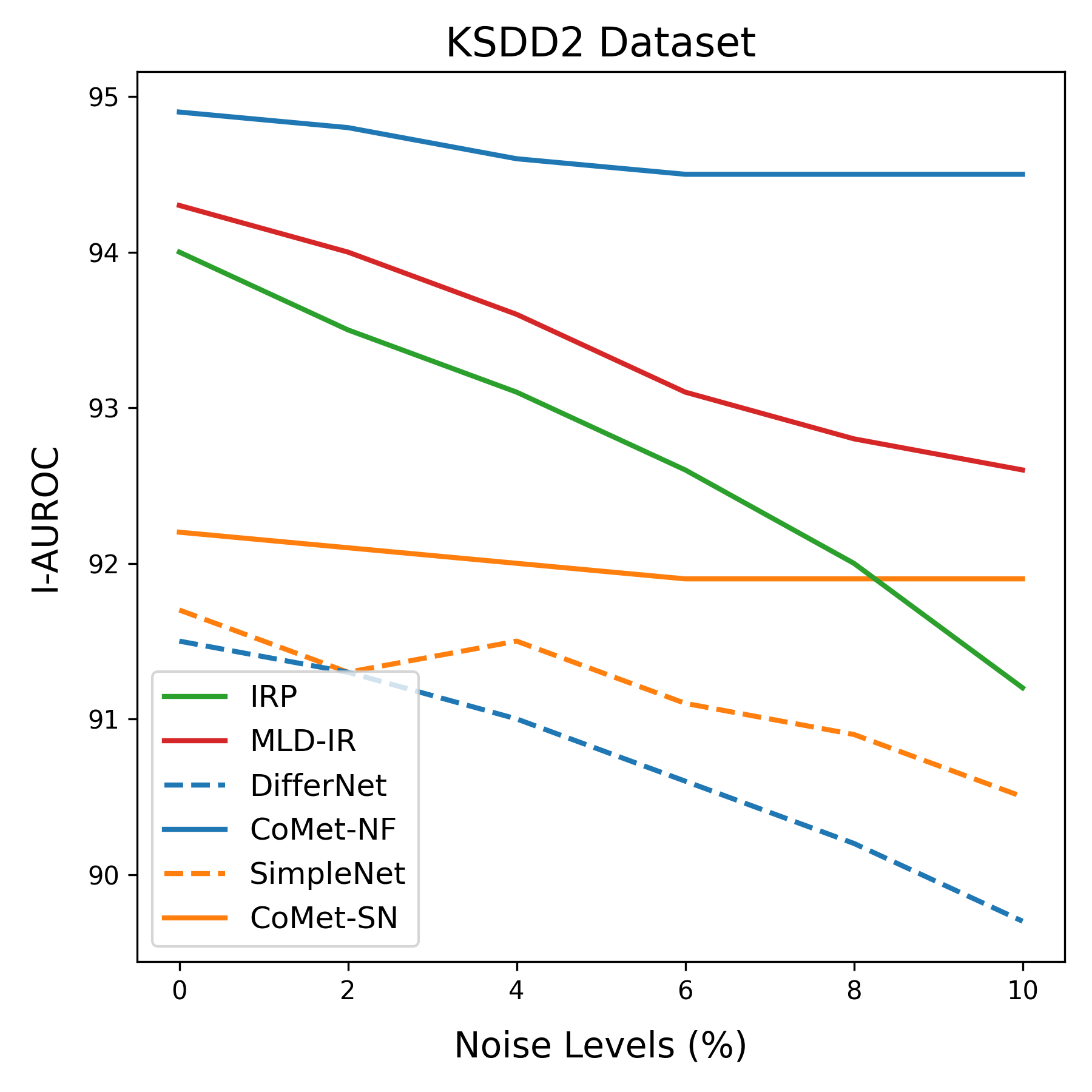}
        \caption{}
        \label{fig:image2}
    \end{subfigure}
    \caption{Comparison of anomaly detection methods on the MVtec-AD (a) and KSDD2 (b) datasets, showing I-AUROC values across noise levels from 0\% to 10\%.}
    \label{fig:noise}
\end{figure}

We finally tested the robustness of the model when subject to increasing levels of noise in the input data. Noise is defined as the number of anomalous samples included in the training set. Figure~\ref{fig:noise} reports I-AUROC performance for noise levels ranging from 0\% to 10\%. We report comparison against both DifferNet and SimpleNet baselines, and two Robust Anomaly Detection methods IRP \cite{aqeel2024selfirp} and MLD-IR \cite{aqeel2025meta} that are specifically designed to address this kind of scenario. Besides being the best performing among the different alternatives, \methname also sustains high AUROC with minimal degradation 
when noise intensity increases up to 10\%. 
This experiment is crucial to prove the suitability of our proposed approach to handle training datasets containing both nominal and anomalous samples, moving from semi-supervised to real unsupervised anomaly detection.

\section{Conclusion}
\label{sec:conclusion}

We targeted the problem of unsupervised anomaly detection, where unlabelled nominal and anomalous samples are available at training time.
We presented \methname, an innovative framework for training anomaly detection models that integrates soft confident learning with meta learning to iteratively refine decision boundaries by dynamically identifying and downweighting ambiguous boundary samples. 
Extensive experiments on industrial datasets demonstrate that \methname achieves state-of-the-art performance, significantly improving recall without compromising precision. Moreover, its ability to operate effectively on noisy, uncurated datasets makes it highly adaptable to real-world industrial applications. These results highlight the potential of \methname to advance unsupervised anomaly detection, paving the way for more reliable and scalable defect detection.

\section*{Acknowledgements}

This study was carried out within the PNRR research activities of the
consortium iNEST (Interconnected North-Est Innovation Ecosystem) funded by the European Union Next-GenerationEU (Piano Nazionale di Ripresa e Resilienza (PNRR) – Missione 4 Componente 2, Investimento 1.5 – D.D. 1058  23/06/2022, ECS\_00000043).
{
    \small
    \bibliographystyle{ieeenat_fullname}
    \bibliography{main}
}

\end{document}


\maketitle

\begin{table}[ht]
    \centering
    \begin{tabular}{lcccc}
        \hline
        Category & \multicolumn{2}{c}{Previous} & \multicolumn{2}{c}{Updated} \\
        & Precision & Recall & Precision & Recall \\
        \hline
        Carpet & 100 & 58.3 & 88.5 & 93.1 \\
        Grid & 94.4 & 37.0 & 86.7 & 91.3 \\
        Leather & 100 & 82.4 & 93.6 & 93.2 \\
        Tile & 100 & 89.6 & 94.8 & 94.0 \\
        Wood & 100 & 91.4 & 97.8 & 95.7 \\
        Bottle & 100 & 89.5 & 98.2 & 91.8 \\
        Cable & 97.9 & 63.5 & 89.5 & 93.2 \\
        Capsule & 90.1 & 83.0 & 90.3 & 93.2 \\
        Hazelnut & 100 & 91.1 & 98.1 & 92.9 \\
        Metalnut & 98.5 & 89.3 & 95.9 & 93.3 \\
        Pill & 94.2 & 71.7 & 91.4 & 93.8 \\
        Screw & 87.0 & 62.5 & 88.2 & 93.8 \\
        Toothbrush & 100 & 93.8 & 99.1 & 95.8 \\
        Transistor & 76.0 & 59.4 & 82.5 & 90.6 \\
        Zipper & 96.4 & 83.3 & 92.6 & 94.8 \\
        \hline
    \end{tabular}
    \caption{Precision Recall Comparison of MVtec AD using \methname-NF.}
    \label{tab:threshold_metrics_trimmed}
\end{table}


\begin{table}[ht]
    \centering
    \begin{tabular}{lcccc}
        \hline
        Category & \multicolumn{2}{c}{Previous} & \multicolumn{2}{c}{Updated} \\
        & Precision & Recall & Precision & Recall \\
        \hline
        Shredded CR. & 95.2 & 71.8 & 87.4 & 92.7 \\
        Encoder & 52.6 & 50.0 & 59.8 & 85.0 \\
        Raspberry & 92.4 & 81.3 & 87.2 & 93.3 \\
        Device Box & 57.1 & 60.0 & 62.7 & 90.0 \\
        L-Fitting & 81.2 & 65.0 & 74.1 & 90.0 \\
        Threaded F. & 58.6 & 97.6 & 58.4 & 88.1 \\
        Redon needle & 93.8 & 77.6 & 91.2 & 93.1 \\
        Aluminium P. & 100.0 & 70.3 & 94.4 & 91.4 \\
        PaperClip & 90.6 & 72.5 & 87.3 & 92.5 \\
        Air ML. & 86.1 & 77.5 & 84.8 & 92.5 \\
        Saw Blade & 62.5 & 50.0 & 60.9 & 90.0 \\
        \hline
    \end{tabular}
    \caption{Precision Recall Comparison of Different Models Without AUROC Values using \methname-NF.}
    \label{tab:precision_recall_comparison_trimmed}
\end{table}


\begin{figure}[t!]
    \centering
    \begin{subfigure}{\linewidth}
        \centering
        \includegraphics[width=\linewidth]{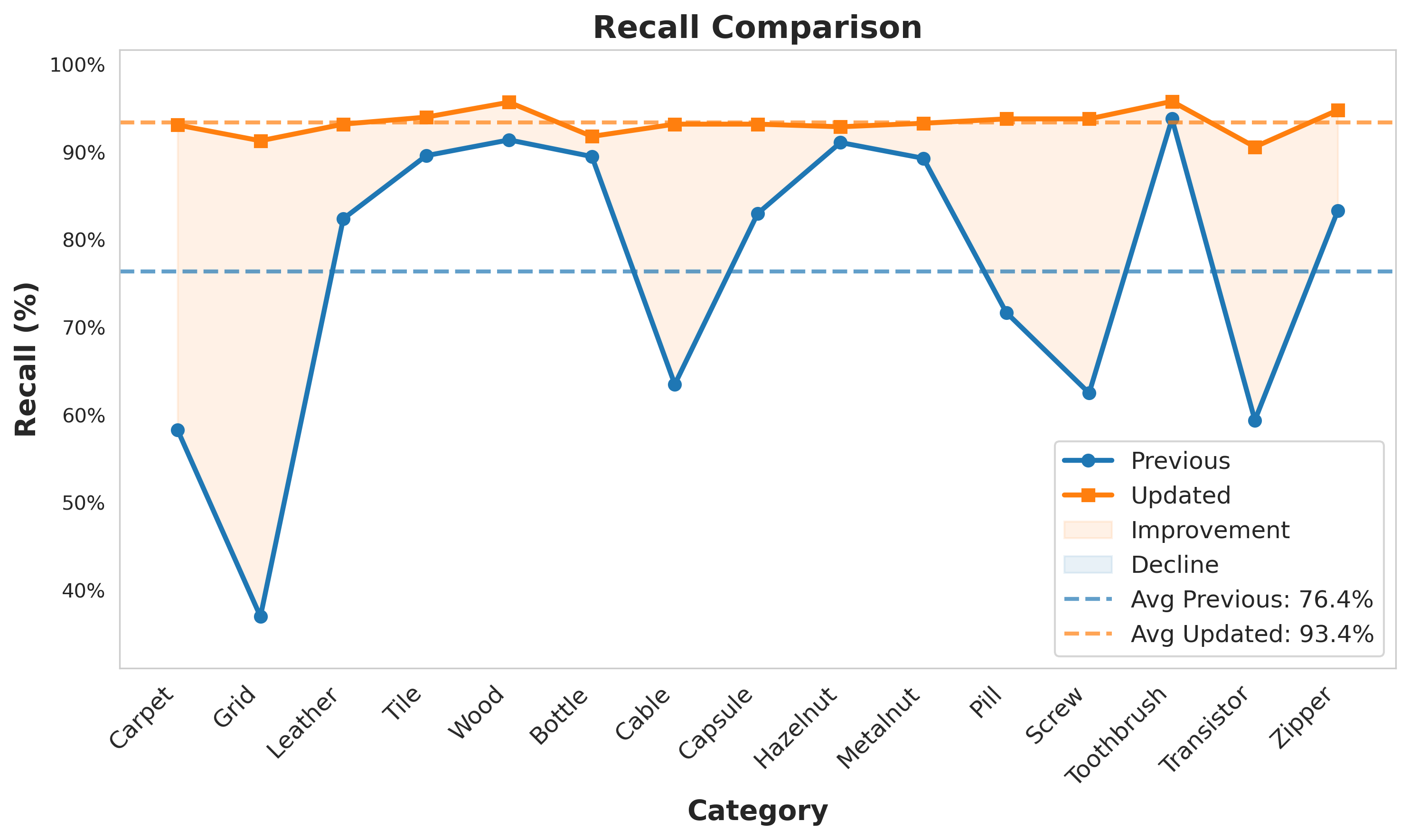}
        \caption{Recall comparison on MVTec.}
        \label{fig:recall_mvtec}
    \end{subfigure}
    
    \vspace{0.5cm} 

    \begin{subfigure}{\linewidth}
        \centering
        \includegraphics[width=\linewidth]{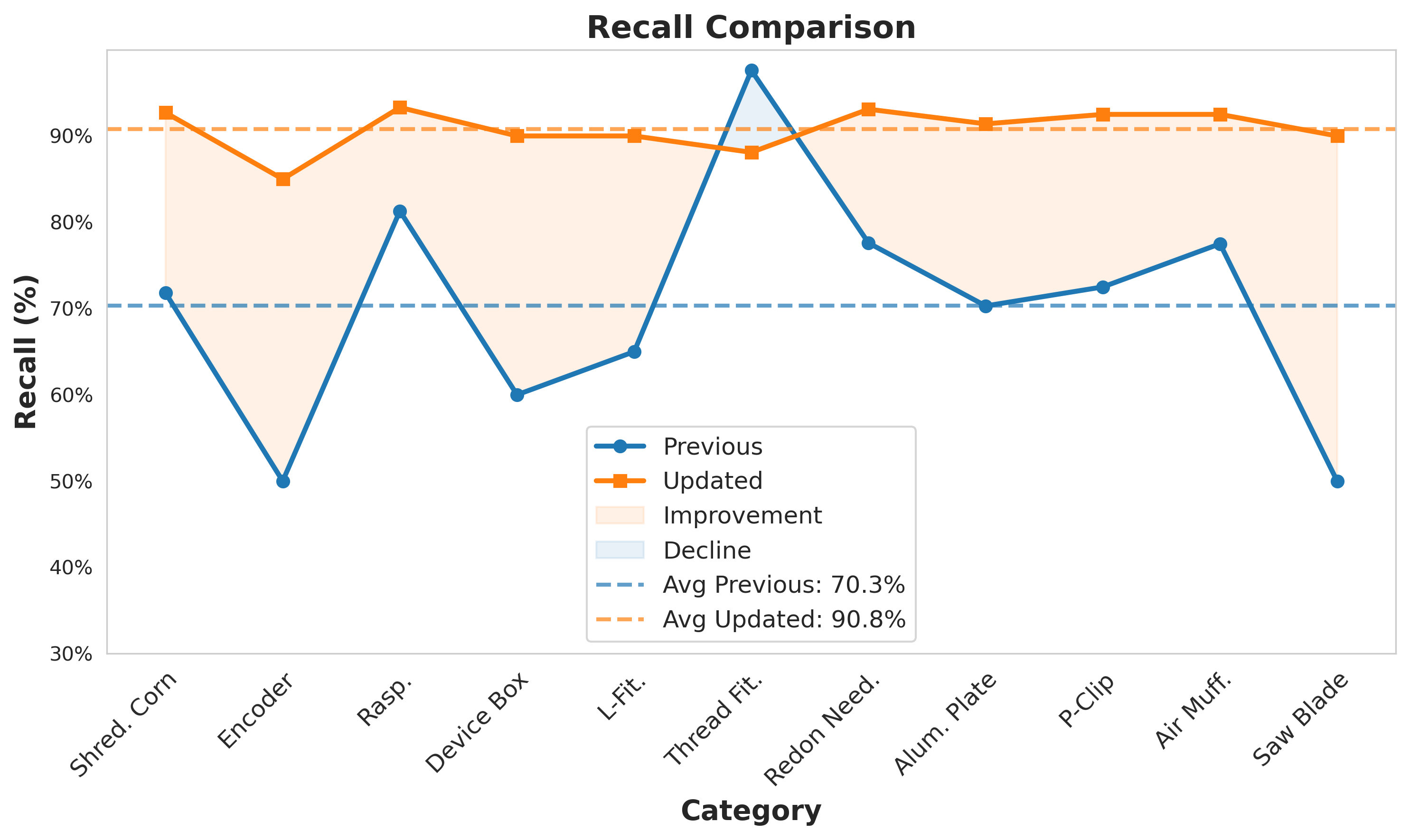}
        \caption{Recall comparison on Viaduct.}
        \label{fig:recall_viaduct}
    \end{subfigure}
    
    \caption{Recall comparison on MVTec and Viaduct datasets.}
    \label{fig:recall_combined}
\end{figure}

\begin{table}[ht]
    \centering
    \begin{tabular}{lcccc}
        \hline
        Dataset & \multicolumn{2}{c}{DifferNet} & \multicolumn{2}{c}{\methname-NF} \\
        & Precision & Recall & Precision & Recall \\
        \hline
        MVTec AD & 95.6 & 76.4 & 92.5 & 93.4 \\
        VIADUCT & 79.1 & 70.3 & 77.1 & 90.8 \\
        KSDD2 & 90.9 & 87.5 & 87.4 & 94.3 \\
        \hline
    \end{tabular}
    \caption{Average Precision and Recall for previous and updated models on MVTec AD, VIADUCT and KSDD2 datasets using \methname-NF.}
    \label{tab:avg-recall}
\end{table}

\begin{table}[!ht]
\centering
\begin{tabular}{lcc|cc}
\toprule
Category & \multicolumn{2}{c|}{SimplNet} & \multicolumn{2}{c}{\methname-SN} \\
 & Precision & Recall & Precision & Recall \\
\midrule
Shredded CR. & 88.5 & 95.2 & 91.3 & 97.8 \\
Encoder & 65.9 & 100 & 78.1 & 86.2 \\
Raspberry & 94.7 & 94.7 & 97.8 & 94.7 \\
Device Box & 92.9 & 84.8 & 91.7 & 95.6 \\
L Fitting & 72.2 & 100 & 86.7 & 100 \\
Threaded Fitting & 72.2 & 62.9 & 69.8 & 88.6 \\
Reedon Needle & 100 & 98.6 & 100 & 100 \\
Aluminium Plate & 84.3 & 97.1 & 89.5 & 98.3 \\
PaperClip & 97.3 & 98.6 & 98.6 & 100 \\
Air Muffler Large & 98.0 & 98.0 & 98.0 & 100 \\
Saw Blade & 60.0 & 100 & 60.0 & 100 \\
\midrule
Average & 84.1 & 93.6 & 87.4 & 96.5 \\
\bottomrule
\end{tabular}
\caption{Comparison of Precision and Recall for on VIADUCT dataset using \methname-SN.}
\label{tab:simplnet_sn}
\end{table}

\begin{table}[!ht]
\centering
\begin{tabular}{lcc|cc}
\toprule
Category & \multicolumn{2}{c|}{SimplNet} & \multicolumn{2}{c}{\methname-SN} \\
 & Precision & Recall & Precision & Recall \\
\midrule
Screw & 94.4 & 95.8 & 97.4 & 98.3 \\
Pill & 97.5 & 95.7 & 96.9 & 98.6 \\
Capsule & 98.2 & 100 & 99.1 & 100 \\
Carpet & 95.6 & 96.6 & 100.0 & 98.9 \\
Grid & 98.3 & 98.2 & 100 & 100 \\
Tile & 97.7 & 98.8 & 100.0 & 100 \\
Wood & 100 & 100 & 100 & 100 \\
Zipper & 99.2 & 100 & 99.2 & 100 \\
Cable & 98.9 & 97.8 & 97.9 & 100 \\
Toothbrush & 96.8 & 100 & 100 & 100 \\
Transistor & 100 & 100 & 100 & 100 \\
Metal Nut & 100 & 100 & 100 & 100 \\
Bottle & 98.4 & 100 & 100 & 100 \\
Hazelnut & 97.2 & 100 & 98.6 & 100 \\
Leather & 100 & 100 & 100 & 100 \\
\midrule
Average & 98.1 & 98.9 & 99.3 & 99.8 \\
\bottomrule
\end{tabular}
\caption{Comparison of Precision and Recall for SimplNet and -SN across multiple categories on MVTec-AD.}
\label{tab:simplnet_sn_extended}
\end{table}

\begin{table*}[ht]
\centering
\resizebox{\textwidth}{!}{%
\begin{tabular}{lcccccccl}
\toprule
\textbf{Method} & \textbf{PatchCore}\cite{roth2022towards} & \textbf{EfficientAD}\cite{batzner2024efficientad} & \textbf{DRÆM}\cite{zavrtanik2021draem} & \textbf{TransFusion}\cite{fuvcka2025transfusion} & \textbf{DiffAD}\cite{zhang2023unsupervised} & \textbf{DifferNet}\cite{RudWan2021} & \textbf{AST}\cite{rudolph2023asymmetric} & \textbf{Our} \\ 
\midrule
\textbf{Carpet}      & 98.7/{\color{gray}\textbf{99.0}} & \textbf{99.3}/- & 97.0/{\color{gray}95.5} & 99.2/- & 98.3/- & 92.9/- & 99.1/- & 98.2/{\color{gray}96.5}\\
\textbf{Grid}        & 98.2/{\color{gray}98.7} & 99.9/- & 99.9/{\color{gray}99.7} & 99.5/- & \textbf{100}/- & 84.0/- & 98.7/- & 97.5/{\color{gray}96.6}\\
\textbf{Leather}     & \textbf{100}/{\color{gray}\textbf{99.3}} & \textbf{100}/- & \textbf{100}/{\color{gray}98.6} & \textbf{100}/- & \textbf{100}/- & 97.1/- & \textbf{100}/- & \textbf{100}/{\color{gray}98.2}\\
\textbf{Tile}        & 98.7/{\color{gray}95.6} & 99.9/- & 99.6/{\color{gray}99.2} & \textbf{100}/- & \textbf{100}/- & 99.4/- & 99.1/- & \textbf{100}/{\color{gray}\textbf{99.2}}\\
\textbf{Wood}        & 99.2/{\color{gray}95.0} & \textbf{100}/- & 99.1/{\color{gray}96.4} & \textbf{100}/- & \textbf{100}/- & 99.8/- & 99.2/- & \textbf{100}/{\color{gray}\textbf{99.8}}\\
\midrule
\textbf{Bottle}      & \textbf{100}/{\color{gray}98.6} & 99.9/- & 99.2/{\color{gray}99.1} & \textbf{100}/- & \textbf{100}/- & 99.0/- & \textbf{100}/- & \textbf{100}/{\color{gray}\textbf{99.8}}\\
\textbf{Cable}       & \textbf{99.5}/{\color{gray}\textbf{98.4}} & 95.2/- & 91.8/{\color{gray}94.7} & 98.5/- & 94.6/- & 95.9/- & 98.5/- & 98.9/{\color{gray}97.6}\\
\textbf{Capsule}     & 98.1/{\color{gray}\textbf{98.8}} & 97.9/- & \textbf{98.5}/{\color{gray}94.3} & 94.0/- & 97.5/- & 86.9/- & 99.7/- & 98.1/{\color{gray}96.4}\\
\textbf{Hazelnut}    & 99.9/{\color{gray}98.7} & 99.4/- & \textbf{100}/{\color{gray}\textbf{99.7}} & \textbf{100}/- & \textbf{100}/- & 99.3/- & \textbf{100}/- & \textbf{100}/{\color{gray}99.6}\\
\textbf{Metal Nut}   & \textbf{100}/{\color{gray}98.4} & 99.6/- & 98.7/{\color{gray}\textbf{99.5}} & \textbf{100}/- & 98.7/- & 96.1/- & 98.5/- & 99.7/{\color{gray}98.9}\\
\textbf{Pill}        & 96.6/{\color{gray}97.4} & 98.6/- & 98.9/{\color{gray}\textbf{97.6}} & 98.3/- & 97.7/- & 88.8/- & \textbf{99.1}/- & 98.6/{\color{gray}96.8}\\
\textbf{Screw}       & 98.1/{\color{gray}\textbf{99.4}} & 96.9/- & 93.9/{\color{gray}97.6} & 86.6/- & 97.2/- & 96.3/- & \textbf{99.7}/- & 98.3/{\color{gray}96.6}\\
\textbf{Toothbrush}  & \textbf{100}/{\color{gray}98.7} & \textbf{100}/- & \textbf{100}/{\color{gray}98.1} & 99.7/- & \textbf{100}/- & 98.6/- & 96.6/- & \textbf{100}/{\color{gray}\textbf{100}}\\
\textbf{Transistor}  & \textbf{100}/{\color{gray}96.3} & 99.9/- & 93.1/{\color{gray}90.9} & 96.6/- & 96.1/- & 91.1/- & 99.3/- & 99.1/{\color{gray}\textbf{98.6}}\\
\textbf{Zipper}      & 99.4/{\color{gray}98.4} & 99.7/- & \textbf{100}/{\color{gray}\textbf{98.8}} & \textbf{100}/- & \textbf{100}/- & 95.1/- & 99.1/- & 99.6/{\color{gray}96.8}\\
\midrule
\textit{Average}     & 99.1/{\color{gray}98.1} & 99.1/- & 98.0/{\color{gray}97.3} & 98.1/- & 98.7/- & 94.9/- & \textbf{99.2}/- & \textbf{99.2}/{\color{gray}\textbf{98.09}}\\ 
\bottomrule
\end{tabular}%
}
\caption{Results of anomaly detection on MVTec-AD dataset (I-AUROC/{\color{gray}P-AUROC} in \%).}
\label{tab:mvtec}
\end{table*}

\begin{table*}[!ht]
\centering
\resizebox{\textwidth}{!}{%
\begin{tabular}{lcccccccc}
\toprule
\textbf{Method} & \textbf{EfficientAD}\cite{batzner2024efficientad} & \textbf{PatchCore}\cite{roth2022towards} & \textbf{MSFlow}\cite{zhou2024msflow} & \textbf{FastFlow}\cite{yu2021fastflow} & \textbf{DRÆM}\cite{zavrtanik2021draem} & \textbf{CFA}\cite{lee2022cfa} & \textbf{MemSeq}\cite{yang2023memseg} & \textbf{Our}\\ 
\midrule
\textbf{Shredded Corn Rough} & 84.1/- & 89.1/- & \textbf{94.1}/- & 74.1/- & 51.4/- & 64.3/- & 82.4/- & 85.2/{\color{gray}88.6}\\
\textbf{Encoder} & 51.2/- & \textbf{70.9}/- & 51.9/- & 52.8/- & 63.0/- & 55.6/- & 59.4/- & 64.3/{\color{gray}61.2}\\
\textbf{Raspberry} & 94.7/- & \textbf{98.9}/- & 92.8/- & 70.7/- & 90.0/- & 89.3/- & 89.9/- & 93.6/{\color{gray}92.2}\\
\textbf{Device Box} & \textbf{87.5}/- & 75.3/- & 79.6/- & 59.4/- & 50.8/- & 56.1/- & 55.7/- & 72.7/{\color{gray}68.4}\\
\textbf{L-Fitting} & \textbf{88.8}/- & 80.6/- & 69.6/- & 55.8/- & 55.7/- & 56.7/- & 66.4/- & 82.4/{\color{gray}80.2}\\
\textbf{Threaded Fitting} & 59.6/- & 55.0/- & 58.6/- & 64.2/- & 60.5/- & 55.1/- & 57.4/- & \textbf{64.9}/{\color{gray}64.7}\\
\textbf{Redon needle} & 98.1/- & 97.2/- & \textbf{100}/- & 69.3/- & 66.8/- & 80.3/- & 51.9/- & \textbf{100}/{\color{gray}95.8}\\
\textbf{Aluminium Plate} & 89.4/- & 86.6/- & \textbf{96.2}/- & 54.7/- & 56.7/- & 50.8/- & 62.9/- & 92.3/{\color{gray}\textbf{93.5}}\\
\textbf{PaperClip} & 97.2/- & \textbf{99.8}/- & 99.2/- & 58.4/- & 69.5/- & 78.3/- & 69.6/- & 97.9/{\color{gray}94.2}\\
\textbf{Air Muffler Large} & 89.1/- & \textbf{99.8}/- & 94.7/- & 58.9/- & 55.4/- & 76.3/- & 80.7/- & 94.5/{\color{gray}91.6}\\
\textbf{Saw Blade} & 69.2/- & 61.0/- & 70.2/- & 51.7/- & 58.2/- & 55.7/- & 56.6/- & \textbf{70.3}/{\color{gray}66.5}\\
\midrule
\textit{Average} & 82.6/- & 83.1/- & 82.4/- & 60.9/- & 61.6/- & 65.3/- & 66.6/- & \textbf{83.5}/{\color{gray}81.5}\\ 
\bottomrule
\end{tabular}%
}
\caption{Results of anomaly detection on VIADUCT dataset (I-AUROC/{\color{gray}P-AUROC} in \%).}
\label{tab:viaduct}
\end{table*}


\begin{table*}[!ht]
\centering
\renewcommand{\arraystretch}{1.2}
\resizebox{\textwidth}{!}{%
\begin{tabular}{cccccccccc}
\hline
\textbf{Method} & \textbf{US}\cite{bergmann2020uninformed} & \textbf{MAD}\cite{rippel2021modeling} & \textbf{MLD-IR}\cite{aqeel2024meta} & \textbf{DRÆM}\cite{zavrtanik2021draem} & \textbf{OSR}\cite{aqeel2024self} & \textbf{PaDim}\cite{defard2021padim} & \textbf{DSR}\cite{zavrtanik2022dsr} & \textbf{IRP}\cite{aqeel2024selfirp} & \textbf{Our} \\
\hline
\textbf{KSDD2} & 65.3/- & 79.3/- & 94.3/- & 77.8/- & 92.1/- & 55.6/- & 87.2/- & 94.0/- & \textbf{94.9}/{\color{gray}91.7} \\
\hline
\end{tabular}%
}
\caption{Performance of Different Methods on KSDD2 Dataset (I-AUROC/{\color{gray}P-AUROC} in \%).}
\label{tab:ksdd2}
\end{table*}

\begin{table*}[ht]
    \centering
    \renewcommand{\arraystretch}{1.2}
    \begin{tabular}{lcccccc}
        \toprule
        \multirow{2}{*}{Category} & \multicolumn{6}{c}{Noise Level} \\
        \cmidrule(lr){2-7}
         & 0\% & 2\% & 4\% & 6\% & 8\% & 10\% \\
        \midrule
        Carpet      & 99.7 / 98.2  & 98.9 / 97.4  & 99.5 / 98.4  & 98.0 / 98.0  & 98.8 / 95.4  & 98.5 / 95.4  \\
        Grid        & 99.7 / 98.8  & 99.7 / 98.5  & 99.8 / 98.4  & 99.9 / 98.8  & 99.6 / 98.8  & 99.3 / 98.8  \\
        Leather     & 100 / 99.2 & 100 / 98.7 & 99.8 / 99.4  & 100 / 99.5 & 99.6 / 99.4  & 99.3 / 99.4  
        \\
        Tile        & 99.8 / 97.0  & 99.9 / 94.5  & 99.7 / 80.5  & 99.9 / 95.1  & 98.4 / 91.1  & 98.1 / 91.1  \\
        Wood        & 100 / 94.5 & 100 / 93.8 & 100 / 94.3 & 100 / 95.1 & 99.3 / 92.4  & 99.0 / 92.4  \\
        \midrule
        Bottle      & 100 / 96.9  & 100.0 / 98.0  & 99.9 / 97.2  & 99.7 / 97.1  & 99.4 / 96.3  & 99.0 / 96.3  \\
        Cable       & 99.9 / 97.6  & 99.8 / 97.3  & 99.9 / 96.2  & 99.4 / 95.9  & 98.0 / 91.9  & 97.7 / 91.9  \\
        Capsule     & 97.7 / 98.9  & 97.0 / 99.0  & 97.4 / 98.8  & 96.1 / 98.6  & 96.4 / 98.8  & 96.1 / 98.8  \\
        Hazelnut    & 100 / 97.9  & 99.5 / 98.0  & 99.3 / 97.8  & 98.9 / 97.5  & 97.9 / 96.1  & 97.5 / 96.1  \\
        Metal Nut   & 100 / 98.8 & 100 / 94.3 & 100 / 93.8 & 100 / 88.8 & 99.5 / 82.5  & 99.1 / 82.5  
        \\
        Pill        & 99.0 / 95.1  & 97.5 / 98.4  & 97.2 / 97.8  & 96.5 / 98.0  & 96.8 / 93.9  & 96.5 / 93.9  \\
        Screw       & 98.2 / 99.3  & 98.8 / 99.1  & 91.8 / 99.2  & 96.5 / 98.7  & 90.8 / 99.0  & 86.4 / 99.0  \\
        Toothbrush  & 99.7 / 98.5  & 99.7 / 98.4  & 99.1 / 98.5  & 98.9 / 98.3  & 99.6 / 98.1  & 99.3 / 98.1  \\
        Transistor  & 100 / 97.6 & 100 / 96.7 & 100 / 96.5 & 100 / 96.6 & 99.6 / 95.2  & 99.3 / 95.2  
        \\
        Zipper      & 99.9 / 98.9  & 99.9 / 98.5  & 99.9 / 97.8  & 99.8 / 97.9  & 98.7 / 95.2  & 98.4 / 95.2  \\
        \midrule
        Average        & 99.6 / 98.1  & 99.4 / 97.4  & 99.1 / 96.3  & 99.0 / 96.9  & 98.8 / 94.9  & 98.5 / 94.9  \\
        \bottomrule
    \end{tabular}
    \caption{Performance results of SimpleNet under different noise levels}
    \label{tab:simplenet1}
\end{table*}

\begin{table*}[ht]
    \centering
    \renewcommand{\arraystretch}{1.2}
    \begin{tabular}{lcccccc}
        \toprule
        \multirow{2}{*}{Category} & \multicolumn{6}{c}{Noise Level} \\
        \cmidrule(lr){2-7}
         & 0\% & 2\% & 4\% & 6\% & 8\% & 10\% \\
        \midrule
        Screw       & 98.7 / 99.8  & 99.0 / 99.6  & 98.5 / 99.2  & 97.8 / 99.0  & 98.0 / 99.2  & 97.9 / 99.2  \\
        Pill        & 98.8 / 98.9  & 98.6 / 99.0  & 98.0 / 98.0  & 97.7 / 98.5  & 97.7 / 97.2  & 98.2 / 92.7  \\
        Capsule     & 98.1 / 99.6  & 97.6 / 99.5  & 97.3 / 98.8  & 97.0 / 98.7  & 96.4 / 98.6  & 96.9 / 98.8  \\
        Carpet      & 99.7 / 98.3  & 99.3 / 98.2  & 100 / 98.9  & 99.9 / 98.3  & 100 / 97.6  & 100 / 99.3  \\
        Grid        & 100 / 99.0  & 99.8 / 99.3  & 99.7 / 98.6  & 100 / 97.9  & 99.8 / 98.5  & 100 / 98.8  \\
        Tile        & 100 / 94.1  & 100 / 94.6  & 99.7 / 87.8  & 99.8 / 96.9  & 99.8 / 91.2  & 99.7 / 95.0  \\
        Wood        & 100 / 94.4  & 100 / 95.2  & 100 / 94.4  & 100 / 95.2  & 100 / 94.8  & 99.8 / 93.3  \\
        Zipper      & 100 / 99.4  & 99.9 / 99.2  & 99.8 / 97.7  & 99.7 / 98.2  & 99.4 / 97.2  & 99.2 / 96.6  \\
        Cable       & 100 / 98.1  & 100 / 98.1  & 99.7 / 95.6  & 99.9 / 96.2  & 99.6 / 95.8  & 99.4 / 93.1  \\
        Toothbrush  & 100 / 99.1  & 100 / 98.8  & 99.7 / 98.5  & 99.7 / 98.5  & 99.7 / 98.7  & 99.7 / 98.1  \\
        Transistor  & 100 / 97.5  & 100 / 97.3  & 100 / 96.5  & 100 / 96.8  & 100 / 96.0  & 100 / 96.0  \\
        Metal Nut   & 100 / 99.4  & 100 / 95.4  & 100 / 94.0  & 100 / 94.4  & 99.9 / 94.9  & 100 / 91.3  \\
        Bottle      & 100 / 98.6  & 100 / 98.7  & 99.9 / 97.7  & 99.9 / 97.9  & 100 / 98.4  & 99.9 / 97.0  \\
        Hazelnut    & 100 / 98.3  & 99.7 / 98.5  & 99.8 / 98.0  & 99.8 / 97.8  & 99.1 / 98.0  & 99.7 / 93.9  \\
        Leather     & 100 / 99.8  & 100 / 99.6  & 100 / 99.6  & 100 / 99.4  & 100 / 99.4  & 100 / 99.5  \\
        \midrule
        Average     & 99.7 / 98.3  & 99.6 / 98.1  & 99.5 / 96.9  & 99.4 / 97.6  & 99.3 / 97.0  & 99.4 / 96.2  \\
        \bottomrule
    \end{tabular}
    \caption{Performance results of different categories under varying noise levels \methname-SN.}
    \label{tab:simplenet2}
\end{table*}

\begin{table*}[!ht]
\centering
\resizebox{\textwidth}{!}{%
\begin{tabular}{lcccccccc}
\toprule
\textbf{Model} & \textbf{DRÆM}\cite{zavrtanik2021draem} & \textbf{CutPaste}\cite{li2021cutpaste} & \textbf{CS-Flow}\cite{rudolph2022fully} & \textbf{PaDiM}\cite{defard2021padim} & \textbf{RevDist}\cite{deng2022anomaly} & \textbf{PatchCore}\cite{roth2022towards} & \textbf{SimpleNet}\cite{liu2023simplenet} & \textbf{Ours} \\ 
\midrule
\textbf{Carpet}      & 97.0/{\color{gray}95.5} & 93.9/{\color{gray}98.3} & \textbf{100}/- & 99.8/{\color{gray}\textbf{99.1}} & 98.9/{\color{gray}98.9} & 98.7/{\color{gray}99.0} & 99.7/{\color{gray}98.2} & 99.7/{\color{gray}98.3} \\
\textbf{Grid}        & 99.9/{\color{gray}97.7} & \textbf{100}/{\color{gray}97.5} & 99.0/- & 96.7/{\color{gray}97.3} & \textbf{100}/{\color{gray}98.9} & 98.2/{\color{gray}98.7} & 99.7/{\color{gray}98.8} & \textbf{100}/{\color{gray}\textbf{99.0}} \\
\textbf{Leather}     & \textbf{100}/{\color{gray}98.6} & \textbf{100}/{\color{gray}99.5} & \textbf{100}/- & \textbf{100}/{\color{gray}99.2} & \textbf{100}/{\color{gray}99.4} & \textbf{100}/{\color{gray}99.3} & \textbf{100}/{\color{gray}99.2} & \textbf{100}/{\color{gray}\textbf{99.8}} \\
\textbf{Tile}        & 99.6/{\color{gray}\textbf{99.2}} & 94.6/{\color{gray}90.5} & \textbf{100}/- & 98.1/{\color{gray}94.1} & 99.3/{\color{gray}95.6} & 98.7/{\color{gray}95.6} & 99.8/{\color{gray}97.0} & \textbf{100}/{\color{gray}94.1} \\
\textbf{Wood}        & 99.1/{\color{gray}\textbf{96.4}} & 99.1/{\color{gray}95.5} & \textbf{100}/- & 99.2/{\color{gray}94.9} & 99.2/{\color{gray}95.3} & 99.2/{\color{gray}95.0} & \textbf{100}/{\color{gray}94.5} & \textbf{100}/{\color{gray}94.4} \\
\midrule
\textbf{Bottle}      & 99.2/{\color{gray}\textbf{99.1}} & 98.2/{\color{gray}97.6} & 99.8/- & 99.1/{\color{gray}98.3} & \textbf{100}/{\color{gray}98.7} & \textbf{100}/{\color{gray}98.6} & \textbf{100}/{\color{gray}96.9} & \textbf{100}/{\color{gray}98.6} \\
\textbf{Cable}       & 91.8/{\color{gray}94.7} & 81.2/{\color{gray}90.0} & 99.1/- & 97.1/{\color{gray}96.7} & 95.0/{\color{gray}97.4} & 99.5/{\color{gray}\textbf{98.4}} & 99.9/{\color{gray}97.6} & \textbf{100}/{\color{gray}98.1} \\
\textbf{Capsule}     & \textbf{98.5}/{\color{gray}94.3} & 98.2/{\color{gray}97.4} & 97.1/- & 87.5/{\color{gray}98.5} & 96.3/{\color{gray}98.7} & 98.1/{\color{gray}98.8} & 97.7/{\color{gray}98.9} & 98.1/{\color{gray}\textbf{99.6}} \\
\textbf{Hazelnut}    & \textbf{100}/{\color{gray}\textbf{99.7}} & 98.3/{\color{gray}97.3} & 99.6/- & 99.4/{\color{gray}98.2} & 99.9/{\color{gray}98.9} & \textbf{100}/{\color{gray}98.7} & \textbf{100}/{\color{gray}97.9} & \textbf{100}/{\color{gray}98.3} \\
\textbf{Metal Nut}   & 98.7/{\color{gray}\textbf{99.5}} & 99.9/{\color{gray}93.1} & 99.1/- & 96.2/{\color{gray}97.2} & \textbf{100}/{\color{gray}97.3} & \textbf{100}/{\color{gray}98.4} & \textbf{100}/{\color{gray}98.8} & \textbf{100}/{\color{gray}99.4} \\
\textbf{Pill}        & 98.9/{\color{gray}97.6} & 94.9/{\color{gray}95.7} & 98.6/- & 90.1/{\color{gray}95.7} & 96.6/{\color{gray}98.3} & 96.6/{\color{gray}97.4} & \textbf{99.0}/{\color{gray}95.1} & 98.8/{\color{gray}\textbf{98.9}} \\
\textbf{Screw}       & 93.9/{\color{gray}97.6} & 88.7/{\color{gray}96.7} & 97.6/- & 97.5/{\color{gray}98.5} & 97.0/{\color{gray}99.6} & 98.1/{\color{gray}99.4} & 98.2/{\color{gray}99.3} & \textbf{98.7}/{\color{gray}\textbf{99.8}} \\
\textbf{Toothbrush}  & \textbf{100}/{\color{gray}98.1} & 99.4/{\color{gray}98.1} & 91.9/- & \textbf{100}/{\color{gray}98.8} & 99.5/{\color{gray}\textbf{99.1}} & \textbf{100}/{\color{gray}98.7} & 99.7/{\color{gray}98.5} & \textbf{100}/{\color{gray}\textbf{99.1}} \\
\textbf{Transistor}  & 93.1/{\color{gray}90.9} & 96.1/{\color{gray}93.0} & 99.3/- & 94.4/{\color{gray}97.5} & 96.7/{\color{gray}92.5} & \textbf{100}/{\color{gray}96.3} & \textbf{100}/{\color{gray}\textbf{97.6}} & \textbf{100}/{\color{gray}97.5} \\
\textbf{Zipper}      & \textbf{100}/{\color{gray}98.8} & 99.9/{\color{gray}99.3} & 99.7/- & 98.6/{\color{gray}98.5} & 98.5/{\color{gray}98.2} & 99.4/{\color{gray}98.8} & 99.9/{\color{gray}98.9} & \textbf{100}/{\color{gray}\textbf{99.4}} \\
\midrule
\textit{Average}     & 98.0/{\color{gray}97.3} & 96.1/{\color{gray}96.0} & 98.7/- & 95.8/{\color{gray}97.5} & 98.5/{\color{gray}97.8} & 99.1/{\color{gray}98.1} & 99.6/{\color{gray}98.1} & \textbf{99.7}/{\color{gray}\textbf{98.3}} \\ 
\bottomrule
\end{tabular}%
}
\caption{Comparison of different models on anomaly detection performance (I-AUROC/{\color{gray}P-AUROC} in \%).}
\label{tab:simplenet3}
\end{table*}

{
    \small
    \bibliographystyle{ieeenat_fullname}
    \bibliography{main}
}